\PassOptionsToPackage{unicode}{hyperref}
\PassOptionsToPackage{hyphens}{url}
\PassOptionsToPackage{dvipsnames,svgnames,x11names}{xcolor}
\documentclass[12pt]{article}

\usepackage{amsmath,amssymb}
\usepackage{iftex}
\ifPDFTeX
  \usepackage[T1]{fontenc}
  \usepackage[utf8]{inputenc}
  \usepackage{textcomp} 
\else 
  \usepackage{unicode-math}
  \defaultfontfeatures{Scale=MatchLowercase}
  \defaultfontfeatures[\rmfamily]{Ligatures=TeX,Scale=1}
\fi
\usepackage{lmodern}
\ifPDFTeX\else  
\fi
\IfFileExists{upquote.sty}{\usepackage{upquote}}{}
\IfFileExists{microtype.sty}{
  \usepackage[]{microtype}
  \UseMicrotypeSet[protrusion]{basicmath} 
}{}
\makeatletter
\@ifundefined{KOMAClassName}{
  \IfFileExists{parskip.sty}{%
    \usepackage{parskip}
  }{
    \setlength{\parindent}{0pt}
    \setlength{\parskip}{6pt plus 2pt minus 1pt}}
}{
  \KOMAoptions{parskip=half}}
\makeatother
\usepackage{xcolor}
\makeatletter
\ifx\paragraph\undefined\else
  \let\oldparagraph\paragraph
  \renewcommand{\paragraph}{
    \@ifstar
      \xxxParagraphStar
      \xxxParagraphNoStar
  }
  \newcommand{\xxxParagraphStar}[1]{\oldparagraph*{#1}\mbox{}}
  \newcommand{\xxxParagraphNoStar}[1]{\oldparagraph{#1}\mbox{}}
\fi
\ifx\subparagraph\undefined\else
  \let\oldsubparagraph\subparagraph
  \renewcommand{\subparagraph}{
    \@ifstar
      \xxxSubParagraphStar
      \xxxSubParagraphNoStar
  }
  \newcommand{\xxxSubParagraphStar}[1]{\oldsubparagraph*{#1}\mbox{}}
  \newcommand{\xxxSubParagraphNoStar}[1]{\oldsubparagraph{#1}\mbox{}}
\fi
\makeatother

\usepackage{longtable,booktabs,array}
\usepackage{calc} 
\usepackage{etoolbox}
\makeatletter
\patchcmd\longtable{\par}{\if@noskipsec\mbox{}\fi\par}{}{}
\makeatother
\IfFileExists{footnotehyper.sty}{\usepackage{footnotehyper}}{\usepackage{footnote}}
\makesavenoteenv{longtable}
\usepackage{graphicx}
\makeatletter
\def\maxwidth{\ifdim\Gin@nat@width>\linewidth\linewidth\else\Gin@nat@width\fi}
\def\maxheight{\ifdim\Gin@nat@height>\textheight\textheight\else\Gin@nat@height\fi}
\makeatother
\setkeys{Gin}{width=\maxwidth,height=\maxheight,keepaspectratio}
\makeatletter
\def\fps@figure{htbp}
\makeatother

\makeatletter
\@ifpackageloaded{caption}{}{\usepackage{caption}}
\AtBeginDocument{%
\ifdefined\contentsname
  \renewcommand*\contentsname{Table of contents}
\else
  \newcommand\contentsname{Table of contents}
\fi
\ifdefined\listfigurename
  \renewcommand*\listfigurename{List of Figures}
\else
  \newcommand\listfigurename{List of Figures}
\fi
\ifdefined\listtablename
  \renewcommand*\listtablename{List of Tables}
\else
  \newcommand\listtablename{List of Tables}
\fi
\ifdefined\figurename
  \renewcommand*\figurename{Figure}
\else
  \newcommand\figurename{Figure}
\fi
\ifdefined\tablename
  \renewcommand*\tablename{Table}
\else
  \newcommand\tablename{Table}
\fi
}
\@ifpackageloaded{float}{}{\usepackage{float}}
\floatstyle{ruled}
\@ifundefined{c@chapter}{\newfloat{codelisting}{h}{lop}}{\newfloat{codelisting}{h}{lop}[chapter]}
\floatname{codelisting}{Listing}

\makeatother
\makeatletter
\@ifpackageloaded{caption}{}{\usepackage{caption}}
\@ifpackageloaded{subcaption}{}{\usepackage{subcaption}}
\makeatother

\ifLuaTeX
  \usepackage{selnolig}  
\fi
\usepackage[]{natbib}
\usepackage{bookmark}

\IfFileExists{xurl.sty}{\usepackage{xurl}}{} 
\hypersetup{
  pdftitle={Conformalized Super Learner},
  pdfauthor={Zhanli Wu, Fabrizio Leisen, Miguel-Angel Luque-Fernandez, F. Javier Rubio},
  pdfkeywords={Conformal prediction, Ensemble, Majority vote, Super Learner, Validity},
  colorlinks=true,
  linkcolor={blue},
  filecolor={Maroon},
  citecolor={Blue},
  urlcolor={Blue},
  pdfcreator={LaTeX via pandoc}}

\newcommand{\anon}{1}

\usepackage{amsthm,amsmath,amsfonts,amssymb}
\usepackage{graphicx}

\usepackage{hyperref}       
\usepackage{url}            
\usepackage{booktabs}       
\usepackage{amsfonts}       
\usepackage{nicefrac}       
\usepackage{microtype}      
\usepackage{graphicx}
\usepackage{caption}
\usepackage{subcaption}
\usepackage{booktabs}
\usepackage{multirow}
\usepackage{amsmath}
\usepackage{amsfonts}
\usepackage{amssymb}
\usepackage{booktabs}
\usepackage{longtable}
\usepackage{algorithm}
\usepackage{algpseudocode}

\usepackage{enumitem}

\usepackage{placeins}

\usepackage{mathtools,upgreek,eucal,mathrsfs,enumitem,amsthm,amsmath}
\usepackage{times,epsfig,makeidx,amsfonts,amsmath,dcolumn,multirow,amssymb,colortbl,url}
\usepackage[flushleft]{threeparttable}

\newcommand{\bxi}{\textbf{\bxi}}

\newcommand{\bw}{{\bf w}}
\newcommand{\bX}{{\bf X}}
\newcommand{\bx}{{\bf x}}

\theoremstyle{plain}

\newtheorem{theorem}{Theorem}[section]

\theoremstyle{definition}

\begin{document}

\def\spacingset#1{\renewcommand{\baselinestretch}%
{#1}\small\normalsize} \spacingset{1}


\if1\anon
{
  \title{\bf Conformalized Super Learner}
  \author{Zhanli Wu\thanks{
    The authors would like to thank the anonymous referees, an Associate
    Editor and the Editor for their constructive comments that improved the
    quality of this paper.
    The first author was supported by NMES Faculty Studentship from the Faculty of Natural, Mathematical and Engineering Sciences, King's College London.}\hspace{.2cm}\\
    Department of Mathematics, King's College London, London, UK\\
    zhanli.wu@kcl.ac.uk\\
    and \\
    Fabrizio Leisen\thanks{Corresponding author.}\\
    Department of Mathematics, King's College London, London, UK\\
    fabrizio.leisen@kcl.ac.uk\\
    and \\
    Miguel-Angel Luque-Fernandez\\
    Department of Statistics and Operations Research, University of Granada, Granada, Spain\\
    mluquefe@ugr.es\\
    and \\
    F. Javier Rubio\\
    Department of Statistical Science, University College London, London, UK\\
    f.j.rubio@ucl.ac.uk}
  \maketitle
} \fi

\if0\anon
{
  \bigskip
  \bigskip
  \bigskip
  \begin{center}
    {\LARGE\bf Conformalized Super Learner}
\end{center}
  \medskip
} \fi

\bigskip
\begin{abstract}
The Super Learner (SL) is a widely used ensemble method that combines point predictions from a library of learners based on their predictive performance. Interval predictions are of considerable practical interest because they allow uncertainty in predictions produced by an individual learner or an ensemble to be quantified. Several methods have been proposed for constructing interval predictions based on the SL, however, these approaches are typically justified using asymptotic arguments or rely on computationally intensive procedures such as the bootstrap.
Conformal prediction (CP) is a machine learning framework for constructing prediction intervals with finite-sample and asymptotic coverage guarantees under mild conditions. We propose coupling CP with the SL through a natural construction that mirrors the original SL framework, using individual learner weights and combining learner-specific conformal prediction sets via a weighted majority vote.
We characterize the properties of the resulting SL-based prediction intervals for continuous outcomes. We cover settings under exchangeability, and data-generating mechanisms exhibiting heteroscedasticity, sparsity, and other forms of distributional heterogeneity. A comprehensive simulation study shows that the conformalized SL achieves valid finite-sample coverage with competitive performance relative to the true data-generating mechanism.
We present an application of this methodology to predict creatinine levels using socio-demographic, biometric, and laboratory measurements. This example demonstrates the benefits of an ensemble with carefully selected learners designed to capture key aspects of complex regression functions, including non-linear effects, interactions, sparsity, heteroscedasticity, and robustness to outliers.
\end{abstract}

\noindent%
\vfill

\newpage
\spacingset{1.65} 

\section{Introduction}\label{sec:intro}

Prediction remains a central objective in theoretical, methodological, and applied research, which is often used to support decision-making in many areas. Both point predictions (for expected or typical outcomes) and interval predictions (for uncertainty quantification and risk assessment) are therefore of primary interest. A wide range of approaches have been proposed for constructing both point and interval predictions, including bootstrap, Monte Carlo, Bayesian, and machine learning approaches. In most cases, however, predictions are generated from a single model or algorithm (learner), such as a regression model or a random forest. This reliance can be limiting in practice, where different learners may capture distinct features of the data, such as heteroscedasticity, sparsity, or non-linear structure. As a result, combining multiple learners offers a principled way to make use of their complementary strengths and improve both point and interval prediction performance.
Stacking (or stacked generalization) is an ensemble learning technique in which multiple base learners are combined to improve predictive performance according to a specified optimality criterion \citep{wolpert:1992,breiman:1996}. More broadly, ensemble methods aim either to select a single learner or to combine their predictions based on predictive performance, or more generally, their performance with respect to a risk function of interest.
A prominent implementation of this principle is the ``Super Learner'' (SL), also known as the stacked ensemble learner \citep{van:2007}. The SL operationalizes stacking by combining predictions from a library of base learners through cross-validation, constructing an optimally weighted ensemble with respect to a user-specified risk function \citep{van:2007,polley:2010b,naimi:2018,phillips:2023}. Under suitable conditions, it is asymptotically guaranteed to perform at least as well as the best candidate learner in the library \citep{van:2006}.
Despite its flexibility and strong theoretical guarantees, the SL produces only point predictions obtained by aggregating the outputs of the ensemble. This represents a limitation, as it does not directly provide a quantification of predictive uncertainty. While several approaches, such as Monte Carlo methods, cross-validation-based procedures, quantile regression, and bootstrap techniques, have been proposed to construct interval predictions \citep{polley:2010,naimi:2018,susmann:2025}, these methods typically rely on asymptotic guarantees and often require relatively strong assumptions.

Conformal prediction (CP; \citealp{vovk:2022}) provides a general framework for constructing prediction intervals with finite-sample (or asymptotic) coverage guarantees under mild conditions, most commonly exchangeability. Its integration with stacking methods has attracted increasing attention, although the existing literature remains heterogeneous in both objectives and assumptions.
Early work by \cite{neeven:2018} considered conformal approaches for stacking weather forecasts derived from both conformal and non-conformal predictors. However, their analysis was primarily empirical, and no formal guarantees were established for the resulting intervals. Subsequent contributions have extended this line of work in several directions. For instance, \cite{jensen:2022,hajihashemi:2024} study combinations of stacking and CP in temporally structured settings, while synergy conformal prediction \citep{gauraha:2021} aggregates predictors through monotonic non-conformity scores, and \cite{rivera:2026} develop approaches based on multivariate non-conformity scores.
A complementary strand of the literature focuses on aggregation schemes with explicit theoretical guarantees. In this direction, \cite{cherubin:2019} and \cite{gasparin:2024,gasparin:2024b} propose stacking strategies based on majority voting with fixed weights and establish coverage guarantees for the resulting conformal predictors. These works are further extended through refinements such as random thresholding and weighted majority voting with data-independent weights.
More broadly, alternative approaches addressing the trade-off between finite-sample coverage and interval efficiency are discussed in \cite{yang:2025}, while \cite{marques:2025} analyze stacked conformal prediction for ensembles satisfying certain symmetry conditions. More recently, several references have focused on optimizing the size of the aggregated conformal prediction intervals. In this line, \cite{hegazy:2026} introduced a stability-based selection framework to maintain marginal coverage guarantees when choosing the smallest available prediction set. \cite{xu:2026} proposed an aggregation strategy based on intersecting the conformal prediction sets in the ensemble whose that minimize empirical set size while guaranteeing asymptotic coverage. \cite{liang:2026} select, from a collection of pretrained models, the one yielding the smallest conformal prediction set, and correct for the resulting selection bias. A comprehensive overview of aggregation methods for conformal prediction sets is provided in Section 10.4 of \cite{angelopoulos:2024}.

In this work, we study the integration of the SL with CP to construct prediction sets for continuous outcomes, with potential extensions to other outcome types. Following the spirit of the
SL, we treat each learner separately, using \emph{ad hoc} non-conformity measures to construct learner-specific conformal prediction intervals. These are then aggregated into a single
prediction set through a weighted majority vote \citep{gasparin:2024,gasparin:2024b}, with the SL weights. This strategy accommodates both full and split conformal approaches.
Previous work has primarily used the SL within CP to estimate nuisance parameters \citep{yang:2024}. We instead use the SL weights to aggregate learner-specific conformal sets for
the outcome of interest. Unlike the fixed-weight majority-vote schemes discussed above, the weights and the conformal sets are here estimated from the same data, which raises the question
of which coverage guarantees can be established. We show that the Split-CSL achieves finite-sample marginal coverage of at least $1-2\alpha$, improving to $1-\alpha$ whenever a single learner receives more than half of the weight, an event determined before observing the test point. For both split and Full-CSL, asymptotic marginal coverage is at least $1-2\alpha$ under training-conditional validity of the base sets, improving to $1-\alpha$ when a single learner, or a coalition of learners with asymptotically equivalent prediction sets, holds a majority of the weight in the limit. Moreover, asymptotic conditional validity at a given covariate value is inherited from the base sets: if each base set is asymptotically conditionally valid, so is the CSL. These results highlight a trade-off between validity and adaptivity, in which the price of adaptivity is limited, and absent when a learner or a coalition dominates. Each learner can be equipped with a conformal procedure guaranteeing marginal coverage of $1-\alpha$. 
Finally, for any weights, the CSL set lies between the intersection and the union of the learner-specific intervals, which links its width to the composition of the library.

The remainder of the paper is organized as follows. Section \ref{sec:SLCP} introduces the SL and CP, and presents the proposed aggregation methodology based on a weighted majority vote for constructing CSL predictions. Section \ref{sec:theory} develops the theoretical results, establishing marginal and conditional validity under general and interpretable conditions. Section \ref{sec:simulation} reports a simulation study illustrating the coverage properties of the proposed CSL across a range of scenarios and data-generating mechanisms. Section \ref{sec:application} presents a case study based on the widely used NHANES data set, focusing on the prediction of creatinine levels from different types of covariates. Finally, Section \ref{sec:discussion} concludes with a discussion of the main contributions and directions for future research. R code and data can be found in the GitHub repository \url{https://github.com/ZWU-001/CSL}.

\section{The Super Learner and Conformal Prediction}\label{sec:SLCP}

\subsection{The Super Learner}\label{subsec:SL}
Let \(\{(\mathbf{X}_i, Y_i)\}_{i=1}^n\) be an independent sample of size \(n\), where \(\mathbf{X}_i \in \mathcal{X} \subseteq \mathbb{R}^p\) denotes the vector of available covariates and \(Y_i \in \mathcal{Y} \subseteq \mathbb{R}\) the response. Our goal is to predict the outcome corresponding to a new covariate value \(\mathbf{X}_{\text{new}}\).
To this end, consider a library of \(K\) candidate learners \(\mathcal{L}_1, \dots, \mathcal{L}_K\). For each \(k \in \{1, \dots, K\}\), let \(f_k\) denote the corresponding population-level prediction function, interpreted as the limiting predictor obtained from \(\mathcal{L}_k\) as \(n \to \infty\). The SL constructs an aggregated predictor as a convex combination of these learners, as described below.

Define the weight space as the \((K-1)\)-simplex
\[
\Delta^{K-1} = \left\{\mathbf{w} = (w_1, \dots, w_K) \in [0,1]^K : w_k \geq 0,\ \sum_{k=1}^K w_k = 1 \right\},
\]
which is compact. For any \(\mathbf{w} \in \Delta^{K-1}\), define the aggregated predictor
\[
f_{\mathbf{w}}(\mathbf{x}) = \sum_{k=1}^K w_k f_k(\mathbf{x}).
\]

Let \(L : \mathcal{Y} \times \mathcal{Y} \to \mathbb{R}_+\) denote a generic loss function. The population risk associated with a weight vector \(\mathbf{w} \in \Delta^{K-1}\) is defined as
\[
R(\mathbf{w}) = \mathbb{E}\left[ L\big(Y, f_{\mathbf{w}}(\mathbf{X})\big) \right],
\]
where the expectation is taken with respect to the true data generating mechanism.
Define the set of optimal weights as
\[
\mathcal{W}^* = \operatorname*{arg\,min}_{\mathbf{w} \in \Delta^{K-1}} R(\mathbf{w}).
\]
The weights that minimize this risk need not be unique; when $R$ is continuous, $\mathcal W^*$ is a non-empty closed subset of $\Delta^{K-1}$. 
Since the squared-error loss is convex in its second argument and $f_{\mathbf w}$ is linear in $\mathbf w$, $R$ is convex on $\Delta^{K-1}$; if, in addition, $R$ is continuous, $\mathcal W^*$ is a non-empty, closed and convex subset of $\Delta^{K-1}$.

In practice, the population risk \(R(\mathbf{w})\) is unknown and must be estimated from the data. The SL uses \(V\)-fold cross-validation to construct an empirical estimate of the risk. Let \(\mathcal{I}_1, \dots, \mathcal{I}_V\) denote a partition of \(\{1, \dots, n\}\) into \(V\) folds. For each fold \(v \in \{1, \dots, V\}\) and learner \(k\), let \(\widehat{f}_k^{(-v)}\) denote the predictor trained on all observations excluding those in \(\mathcal{I}_v\). Define the cross-validated predictor
\[
\widehat{f}_{\mathbf{w}}^{(-v)}(\mathbf{X}_i) = \sum_{k=1}^K w_k \widehat{f}_k^{(-v)}(\mathbf{X}_i).
\]

The cross-validated empirical risk is then given by
\[
R_n(\mathbf{w}) = \frac{1}{n} \sum_{v=1}^V \sum_{i \in \mathcal{I}_v} 
L\big(Y_i, \widehat{f}_{\mathbf{w}}^{(-v)}(\mathbf{X}_i)\big).
\]

The SL estimator of the weight vector is defined as
\[
\widehat{\mathbf{w}}_n  \in \operatorname*{arg\,min}_{\mathbf{w} \in \Delta^{K-1}} R_n(\mathbf{w}).
\]

A particularly important choice in practice is the squared-error loss
\[
L(y, \widehat{y}) = (y - \widehat{y})^2,
\]
which leads to the usual least-squares formulation of the SL.

Under suitable regularity conditions, the cross-validated risk converges uniformly to the population risk \citep{van:2003,van:2007}. Consequently, the estimator \(\widehat{\mathbf{w}}_n\) converges in probability to the set of optimal weights \(\mathcal{W}^*\). Under the squared-error loss considered here, the SL is asymptotically optimal in the sense that its risk is no greater than that of the best candidate learner or any convex combination thereof, with respect to mean squared error. More generally, the SL framework accommodates a wide class of loss functions (\textit{e.g.}, absolute error or deviance loss for binary outcomes), allowing the notion of optimality to be defined relative to the corresponding risk function. Once the weights \(\widehat{\mathbf{w}}_n\) have been estimated, each learner is refitted on the full data set to obtain predictors \(\widehat{f}_1^{(n)}, \dots, \widehat{f}_K^{(n)}\). The resulting SL predictor for a new covariate vector \(\mathbf{X}_{\text{new}} \in \mathcal{X}\) is given by
\[
\widehat{f}_{\text{SL}}^{(n)}(\mathbf{X}_{\text{new}}) = \sum_{k=1}^K \widehat{w}_{n,k}\,\widehat{f}_k^{(n)}(\mathbf{X}_{\text{new}}),
\]
where \(\widehat{w}_{n,k}\) denotes the \(k\)-th component of the estimated weight vector \(\widehat{\mathbf{w}}_n\) \citep{van:2007,polley:2010}.

\subsection{Conformal prediction}\label{subsec:CP}

Conformal prediction (CP) is a distribution-free framework for constructing prediction sets with finite-sample marginal coverage guarantees, requiring only that the data are exchangeable \citep{shafer:2008,papadopoulos:2008,vovk:2022}. Using the notation introduced in Section \ref{subsec:SL}, let \((\bX_{\text{new}},Y_{\text{new}})\) denote a future observation. Assume that \((\bX_1,Y_1),\dots,(\bX_n,Y_n),(\bX_{\text{new}},Y_{\text{new}})\) are exchangeable, although extensions to non-exchangeable data exist \citep{angelopoulos:2024}, and this condition is satisfied in particular when the observations are independent and identically distributed. A non-conformity measure \(s:\mathcal{X}\times\mathcal{Y}\to\mathbb{R}\) assigns a score to each observation to reflect how well $(\bX_i, Y_i)$ conforms to a model fitted on the data set. 
For a candidate value \(y\in\mathcal{Y}\) of \(Y_{\text{new}}\), the same non-conformity measure is used to evaluate the augmented observation \((\bX_{\text{new}},y)\). The conformal prediction set is then defined as the collection of all candidate values \(y\) whose associated score does not exceed an empirical upper quantile of the non-conformity scores computed from the observed sample, where the quantile level is determined by the chosen value of miscoverage level \(\alpha\in(0,1)\). In this way, CP produces a prediction set \(\mathcal{C}(\bX_{\text{new}})\) satisfying, for each $n$,
\[
\Pr\big(Y_{\text{new}}\in \mathcal{C}(\bX_{\text{new}})\big)\geq 1-\alpha.
\]
The two most common implementations are split CP and full CP. 
The central idea is to assess how unusual a candidate response value is relative to the observed data through a non-conformity measure, retaining those candidate values whose non-conformity score does not exceed an appropriate threshold.
In full CP, this threshold is computed by refitting the model for each candidate value, which is computationally expensive in practice. Alternatively, split CP \citep{papadopoulos:2008}, partitions $\mathcal{D}_n$ into a training set and a held-out calibration set, used to determine the threshold and making it computationally efficient. Full CP uses the full data set, typically leading to shorter prediction intervals, but at a substantially higher computational cost \citep{vovk:2022,angelopoulos:2024}.

\subsection{Prediction interval aggregation}\label{subsec:CP_aggregation}

Consider a library of $K$ learners in a SL framework, each equipped with a corresponding non-conformity measure $s_k$. For a new covariate vector $\bX_{\text{new}}$ and a candidate response value $y$, let $\mathcal{C}_k := \mathcal{C}_k(\bX_{\text{new}})$ denote the conformal prediction interval associated with the $k$-th learner under a chosen miscoverage level $\alpha$, and let $\mathbf{1}_{A_k} := \mathbf{1}_{\{ y \in \mathcal{C}_k(\bX_{\text{new}}) \}}$ denote the corresponding inclusion indicator. Note that $\mathbf{1}_{A_k}$ depends implicitly on $y$. We define an aggregation function $\varphi : \{0,1\}^K \to \{0,1\}$, which maps $(\mathbf{1}_{A_1}, \ldots, \mathbf{1}_{A_K})$ to a single inclusion decision, determining whether $y$ is included in the aggregated prediction set based on the inclusion decisions of the $K$ learners.

Several choices for $\varphi$ are possible. For example, $\varphi(\mathbf{1}_{A_1}, \ldots, \mathbf{1}_{A_K})  = \min\{\mathbf{1}_{A_1}, \ldots, \mathbf{1}_{A_K}\}$,
corresponding to the intersection of the $K$ prediction intervals, and $\varphi(\mathbf{1}_{A_1}, \ldots, \mathbf{1}_{A_K}) 
    = \max\{\mathbf{1}_{A_1}, \ldots, \mathbf{1}_{A_K}\}$, corresponding to their union. Another commonly used choice is the weighted majority vote,
\begin{equation}\label{eq:WMV}
    \varphi(\mathbf{1}_{A_1}, \ldots, \mathbf{1}_{A_K})
    = \mathbf{1}_{\left\{ \sum_{k=1}^{K} w_k \mathbf{1}_{A_k} 
    > \frac{1}{2} \right\}},
    \qquad \mathbf{w} = (w_1, \ldots, w_K) \in \Delta^{K-1}.
\end{equation}
The special case $w_k = 1/K$ for all $k$ corresponds to the equal-weight majority vote.
The weighted majority vote (WMV) provides a compromise between the intersection and union aggregation rules. The intersection is sensitive to anti-conservative learners, as a single 
narrow prediction interval can shrink the combined set and destroy coverage; conversely, the union is sensitive to overly conservative learners, expanding to include the widest interval in the library and potentially producing uninformative prediction sets. The WMV mitigates both extremes by weighting the contribution of each learner according to its estimated performance, so that poorly performing learners, whether too conservative or too anti-conservative, induce limited influence on the combined prediction set. These properties are illustrated through theoretical results and simulation studies in the following sections.
The threshold $1/2$ in the weighted majority vote corresponds to a natural majority decision rule under symmetric aggregation of expert opinions, but other values can also be used leading to super-majority ($>1/2$) or super-minority ($<1/2$) strategies. Interpreting the weights $w_k$ as SL weights, they reflect each learner's relative contribution to the optimal convex combination predictor under a chosen loss function, and thus capture predictive performance in terms of out-of-sample risk within the ensemble. The quantity $\sum_{k=1}^K w_k \mathbf{1}_{A_k}$ measures the total support for inclusion, and the threshold $1/2$ selects the point at which inclusion support dominates exclusion in a symmetric decision problem. Choosing a different threshold would introduce an implicit asymmetry, shifting the rule toward more intersection-like or more union-like behavior.
A limiting case of the WMV is the winner-takes-all (WTA) aggregation rule, which assigns all weight to a single learner in the library, reducing the ensemble to an individual predictor. Let $k^\star = \arg\max_{k \in \{1,\dots,K\}} w_k$. This rule then uses only learner $k^\star$, producing the aggregated conformal set $\mathcal{C}_{k^\star}(\mathbf{X})$, and the corresponding inclusion rule $\varphi(\mathbf{1}_{A_1},\ldots,\mathbf{1}_{A_K}) = \mathbf{1}_{A_{k^\star}}$.

\subsection{Conformalized Super Learning}
We now summarize the proposed Conformalized Super Learner (CSL) strategy for constructing prediction sets by combining conformal prediction intervals from a library of $K$ base learners through SL weights, under both split and full conformal settings. Let us also denote by $(\bx_i,y_i)$ and $\bx_{\text{new}}$ the observed sample and the realized covariate vector of the new point.

To construct prediction intervals within the SL framework while preserving its structure and advantages, we focus on the weighted majority vote as an aggregation strategy, with data-dependent weights estimated in the SL. This aligns with the convex (continuous) SL \citep{polley:2010,polley:2011}, which combines point predictions via convex combinations of the individual learners, as described in Section \ref{subsec:SL}. As an alternative, one may consider the discrete SL, which selects the single learner with the highest weight \citep{phillips:2023}, corresponding to a WTA aggregation strategy; however, this is often less appealing, as it fails to exploit the complementary strengths of the learners in the library when no single learner dominates.
Algorithms \ref{alg:splitCSL} and \ref{alg:fullCSL} present the split and Full-CSL procedures in general terms. In both cases, the SL weights are first estimated based on cross-validated predictive performance. In the Split-CSL, the data set is partitioned into a training set and a calibration set: the training set is used to estimate the SL weights and to fit each base learner, while the calibration set is used to compute the non-conformity scores and calibrate the learner-specific conformal intervals. For the Full-CSL, no sample splitting is introduced. Instead, the full data set is used to estimate the SL weights, and for each candidate response value $y$, each base learner is refitted on the corresponding augmented data set obtained by $(y,\bx_{\text{new}})$ and the full data set. The resulting learner-specific conformal intervals are then combined through the weighted majority vote aggregation based on the SL weights. Algorithms \ref{alg:splitCSL} and \ref{alg:fullCSL} detail the steps to obtain interval predictions based on the proposed CSL.

\section{Marginal and conditional validity}\label{sec:theory}
In this section, we provide theoretical guarantees for the CSL methodology proposed in Section \ref{sec:SLCP}. Establishing these results requires several considerations, as a SL library may include a wide range of learners. 
Let $\mathcal D_n$ denote the observed data and let $(\bX,Y)$ denote a generic test pair. For each learner $\mathcal L_k$, $\mathcal C_k(\cdot\,;\mathcal D_n)$ denotes its conformal prediction set at miscoverage level $\alpha$, and we write
\[
A_{n,k}^C := \big\{Y\notin\mathcal C_k(\bX;\mathcal D_n)\big\},\qquad
P_{n,k} := \Pr\big(A_{n,k}^C \mid \mathcal D_n\big),
\]
for the miscoverage event of learner $\mathcal L_k$ and its training-conditional miscoverage probability. For the Split-CSL, $\mathcal D_n=\mathcal D_{\mathrm{tr}}\cup\mathcal D_{\mathrm{cal}}$, with $n_{\mathrm{tr}}$ and $n_{\mathrm{cal}}$ observations, respectively. The SL weights $\widehat{\bw}_n$ and the fitted learners $\widehat f_k$ are functions of $\mathcal D_{\mathrm{tr}}$ only (the $V$-fold cross-validation of the SL is performed within $\mathcal D_{\mathrm{tr}}$), while $\mathcal D_{\mathrm{cal}}$ is used only to calibrate the learner-specific sets. For the Full-CSL, $\widehat{\bw}_n$ is computed from $\mathcal D_n$ by $V$-fold cross-validation,
as described in Section~\ref{subsec:SL}. Throughout, conditioning on $\mathcal D_{\mathrm{tr}}$ or on $\mathcal D_n$ also includes conditioning on the split and cross-validation randomization, which is independent of the test pair. Asymptotic statements refer to $n\to\infty$ with $n_{\mathrm{tr}}\to\infty$ and $n_{\mathrm{cal}}\to\infty$ (for instance, under a fixed split proportion). Finally, we denote the weighted miscoverage vote by
\[
V_n^{\mathrm{mis}} := \sum_{k=1}^K\widehat w_{n,k}\,\mathbf 1_{A_{n,k}^C}.
\]
Under the weighted majority vote (A3 below), $Y\notin\mathcal C_{\mathrm{comb}}(\bX;\mathcal D_n)$ if
and only if $V_n^{\mathrm{mis}}\ge 1/2$. Moreover, at most one weight can exceed $1/2$, and if
$\widehat w_{n,k}>1/2$, then $\mathcal C_{\mathrm{comb}}(\bX;\mathcal D_n)=\mathcal C_k(\bX;\mathcal D_n)$,
since the vote total is at least $\widehat w_{n,k}>1/2$ for $y\in\mathcal C_k(\bX;\mathcal D_n)$ and at
most $1-\widehat w_{n,k}<1/2$ otherwise.

We consider the following conditions.

\begin{itemize}

\item[A0.] The observations $(\bX_1,Y_1),\dots,(\bX_n,Y_n)$ and the test pair $(\bX,Y)$ are exchangeable, any randomization used by the algorithm (the training/calibration split, the
cross-validation folds and any internal randomization of the learners) is independent of the data, and, for the Full-CSL, each learner treats the observations of the augmented data set
symmetrically, so that permuting them leaves the fitted predictor unchanged.

\item[A1.] \emph{(Asymptotic marginal validity of the base sets.)} For each $k\in\{1,\dots,K\}$,
\begin{equation*}
\limsup_{n\to\infty}\Pr\big(Y\notin\mathcal C_k(\bX;\mathcal D_n)\big)\le\alpha .
\end{equation*}

\item[A1$^\star$.] \emph{(Asymptotic training-conditional validity of the base sets.)} For each $k\in\{1,\dots,K\}$ and every $\varepsilon>0$,
\begin{equation*}
\Pr\big(P_{n,k}>\alpha+\varepsilon\big)\to 0,\quad\text{as } n\to\infty .
\end{equation*}

\item[A2.] The estimated weight vector $\widehat{\mathbf{w}}_n$ converges in probability to a closed, convex set of optimal weights $\mathcal{W}^* \subset \Delta^{K-1}$ (the simplex), such that:
\begin{equation*}
\inf_{\mathbf{w} \in \mathcal{W}^*} \|\widehat{\mathbf{w}}_n - \mathbf{w}\| \xrightarrow{\Pr} 0, \quad \text{as } n \to \infty.
\end{equation*}

\item[A3.] The prediction interval is obtained via a weighted majority vote:
\begin{equation*}
\mathcal C_{\text{comb}}(\bX; \mathcal{D}_n)
:= 
\left\{
y : \sum_{k=1}^K \widehat{w}_{n,k} \,  \, \mathbf{1}_{\{y \in \mathcal C_k(\mathbf{X}; \mathcal{D}_n)\}} > \frac{1}{2}
\right\},
\end{equation*}
where the weights satisfy \(\widehat{w}_{n,k} \ge 0\) for each $n$ and \(\sum_{k=1}^K \widehat{w}_{n,k} = 1\). 
\end{itemize}

Condition A0 is used only for the finite-sample result for the Split-CSL (Theorem~\ref{th:splitvalidity}); the asymptotic results do not assume it and use A1 or A1$^\star$
instead. Under A0, split and full conformal sets satisfy A1 for every $n$, while A1 also allows base sets that are only asymptotically valid. Assumption A1$^\star$ strengthens A1: for every
$\varepsilon>0$, the training-conditional miscoverage probability of each base set may exceed $\alpha+\varepsilon$ only with vanishing probability.
Assumption A2 requires only that the distance between $\widehat{\mathbf w}_n$ and the set $\mathcal W^*$ vanishes in probability. The weights need not converge to a point, which accommodates non-unique optimal weights. The SL oracle results \citep{van:2007,polley:2010,polley:2011} concern the risk of the SL rather than the convergence of its weights; Theorem~\ref{theorem:set_consistency} gives sufficient conditions for the latter. Assumption A3 restricts the combined prediction set to a weighted majority vote, as in
Section~\ref{sec:SLCP}. 

The following theorem provides sufficient (but not necessary) conditions on the risk function under which Assumption A2 holds. These conditions are standard in M-estimation \citep{vaart:2000} and related analyses of the SL \citep{li:2025}.

\begin{theorem}[Set-Consistency of Super Learner Weights]\label{theorem:set_consistency}
Let $\Delta^{K-1} = \{ \mathbf{w} \in [0,1]^K : w_k \geq 0, \sum_{k=1}^K w_k = 1\}$ be the $(K-1)$-simplex. Let $R_n(\mathbf{w})$ be the empirical risk (\textit{e.g.}, cross-validated loss) and $R(\mathbf{w})$ be the population risk, which are based on a continuous loss function. Define the set of optimal weights as:
\begin{equation*}
    \mathcal{W}^* = \operatorname{argmin}_{\mathbf{w} \in \Delta^{K-1}} R(\mathbf{w})
\end{equation*}
Suppose the following conditions hold:
\begin{enumerate}
    \item \textbf{Uniform Convergence:} The empirical risk converges uniformly in probability to the population risk:
    \begin{equation*}
        \sup_{\mathbf{w} \in \Delta^{K-1}} |R_n(\mathbf{w}) - R(\mathbf{w})| \xrightarrow{\Pr} 0, \quad \text{as } n \to \infty.
    \end{equation*}
    \item \textbf{Continuity and Well-Separateness:} The population risk function $R(\mathbf{w})$ is continuous on the compact set $\Delta^{K-1}$, and for every $\epsilon > 0$:
    \begin{equation*}
        \inf_{\mathbf{v} : d(\mathbf{v}, \mathcal{W}^*) \geq \epsilon} R(\mathbf{v}) > \min_{\mathbf{w} \in \Delta^{K-1}} R(\mathbf{w})
    \end{equation*}
    where $d(\mathbf{v}, \mathcal{W}^*) = \inf_{\mathbf{w} \in \mathcal{W}^*} \|\mathbf{v} - \mathbf{w}\|$.
\end{enumerate}
Then, any sequence of estimators $\widehat{\mathbf{w}}_n$ such that $R_n(\widehat{\mathbf{w}}_n) \leq \inf_{\mathbf{w} \in \Delta^{K-1}} R_n(\mathbf{w}) + o_p(1)$ satisfies:
\begin{equation*}
    \inf_{\mathbf{w} \in \mathcal{W}^*} \|\widehat{\mathbf{w}}_n - \mathbf{w}\| \xrightarrow{\Pr} 0, \quad \text{as } n \to \infty.
\end{equation*}
\end{theorem}

The following result gives finite-sample lower bounds on the marginal coverage of the Split-CSL: a worst-case bound of $1-2\alpha$, which improves to $1-\alpha$ on the event that a single learner holds more than half of the weight, and a bound for coalitions of learners that depends on the disagreement among their members.
\begin{theorem}[Finite-sample marginal validity of the Split-CSL]\label{th:splitvalidity}
Consider the Split-CSL (Algorithm~\ref{alg:splitCSL}) under A0 and A3, and let $\Omega_n:=\{\max_k\widehat w_{n,k}>1/2\}$ denote the event that a dominant learner exists. Then:
\begin{enumerate}
\item[(a)] \textbf{General case.} For every $n$, $\Pr\big(Y\notin\mathcal C_{\mathrm{comb}}(\bX;\mathcal D_n)\mid\mathcal D_{\mathrm{tr}}\big)\le2\alpha$ almost surely. Hence
\[
\Pr\big(Y\in\mathcal C_{\mathrm{comb}}(\bX;\mathcal D_n)\big)\ge1-2\alpha .
\]
\item[(b)] \textbf{Dominant learner.} For every $n$, $\Pr\big(Y\notin\mathcal C_{\mathrm{comb}}(\bX;\mathcal D_n)\mid\mathcal D_{\mathrm{tr}}\big)\le\alpha$ almost surely on $\Omega_n$. Hence
\[
\Pr\big(Y\in\mathcal C_{\mathrm{comb}}(\bX;\mathcal D_n)\big)\ge 1-\alpha-\alpha\Pr\big(\Omega_n^C\big).
\]
If, in addition, the weights satisfy A2 and there exists $k^\star$ such that $\inf_{\mathbf w\in\mathcal W^*}w_{k^\star}>1/2$, then $\Pr(\Omega_n^C)\to0$ and, consequently,
\[
\liminf_{n\to\infty}\Pr\big(Y\in\mathcal C_{\mathrm{comb}}(\bX;\mathcal D_n)\big)\ge1-\alpha .
\]

\item[(c)] \textbf{Coalition.} Let $\widehat{\mathcal K}\subseteq\{1,\dots,K\}$ be a coalition of learners determined by $\mathcal D_{\mathrm{tr}}$, let
$\Omega_n^{\widehat{\mathcal K}}:=\{\sum_{k\in\widehat{\mathcal K}}\widehat w_{n,k}>1/2\}$, and let
\[
\widehat{\zeta}_n:=\min_{j\in\widehat{\mathcal K}}\sum_{k\in\widehat{\mathcal K}}
\Pr\big(A^C_{n,j}\,\triangle\,A^C_{n,k}\mid\mathcal D_{\mathrm{tr}}\big),
\]
where $\triangle$ denotes the symmetric difference, so that $A^C_{n,j}\triangle A^C_{n,k}$ is the event that exactly one of the learners $j$ and $k$ fails to cover $Y$. Then, for every $n$,
$\Pr\big(Y\notin\mathcal C_{\mathrm{comb}}(\bX;\mathcal D_n)\mid\mathcal D_{\mathrm{tr}}\big) \le\alpha+\min(\widehat{\zeta}_n,\alpha)$ almost surely on $\Omega_n^{\widehat{\mathcal K}}$, and hence
\[
\Pr\big(Y\in\mathcal C_{\mathrm{comb}}(\bX;\mathcal D_n)\big)\ge1-\alpha
-\mathbb E\big[\min(\widehat{\zeta}_n,\alpha)\mathbf 1_{\Omega_n^{\widehat{\mathcal K}}}\big]
-\alpha\Pr\big((\Omega_n^{\widehat{\mathcal K}})^C\big).
\]
If, in addition, $\Pr(\widehat{\mathcal K}=\mathcal K^*)\to1$ for a fixed $\mathcal K^*$ with $\inf_{\mathbf w\in\mathcal W^*}\sum_{k\in\mathcal K^*}w_k\ge\frac12+\delta$ for some $\delta>0$, A2 holds, and $\Pr\big(A^C_{n,i}\triangle A^C_{n,k}\big)\to0$ for all $i,k\in\mathcal K^*$, then $\liminf_{n\to\infty}\Pr\big(Y\in\mathcal C_{\mathrm{comb}}(\bX;\mathcal D_n)\big)\ge1-\alpha$.

\end{enumerate}
\end{theorem}
The bound in part~(c) subtracts from $1-\alpha$ two penalties whose sum is at most $\alpha$: one for the disagreement $\widehat{\zeta}_n$ among coalition members when the coalition holds the
majority, and one for the event that it does not, where only part~(a) applies. A single dominant learner gives $\widehat{\zeta}_n=0$ and recovers part~(b); larger coalitions hold the majority more often but disagree more.
For the Full-CSL, the weights and the conformal sets are computed from the same data, so the argument of Theorem~\ref{th:splitvalidity} does not apply: marginal validity of each base set does not by itself bound the weighted miscoverage, as the weights may be larger for learners whose sets under-cover for the realized data. This idea motivates the following result.

\begin{theorem}[Asymptotic marginal validity]\label{th:marginalvalidity}
Consider the split or Full-CSL under A3.
\begin{enumerate}
\item[(a)] \textbf{General case.} Under A1$^\star$,
\[
\liminf_{n\to\infty}\Pr\big(Y\in\mathcal C_{\mathrm{comb}}(\bX;\mathcal D_n)\big)\ge1-2\alpha .
\]
\item[(b)] \textbf{Dominant learner.} Under A1 and A2, if there exists $k^\star$ such that $\inf_{\mathbf w\in\mathcal W^*}w_{k^\star}>1/2$, then
\[
\liminf_{n\to\infty}\Pr\big(Y\in\mathcal C_{\mathrm{comb}}(\bX;\mathcal D_n)\big)\ge1-\alpha .
\]
\item[(c)] \textbf{Coalition.} Assume A1 and A2, and suppose there exist a fixed subset $\mathcal K^*\subseteq\{1,\dots,K\}$ and $\delta>0$ such that
\[
\inf_{\mathbf w\in\mathcal W^*}\sum_{k\in\mathcal K^*}w_k\ge\tfrac12+\delta
\qquad\text{and}\qquad
\Pr\big(A^C_{n,j}\,\triangle\,A^C_{n,k}\big)\to0\quad\text{for all } j,k\in\mathcal K^*.
\]
Then
\[
\liminf_{n\to\infty}\Pr\big(Y\in\mathcal C_{\mathrm{comb}}(\bX;\mathcal D_n)\big)\ge1-\alpha .
\]
In particular, this holds with $\mathcal K^*=\{k^\star\}$ whenever $\inf_{\mathbf w\in\mathcal W^*}w_{k^\star}>1/2$.
\end{enumerate}
\end{theorem}

The following result extends these guarantees to conditional coverage at a new covariate value.
Distribution-free finite-sample conditional validity is unachievable \citep{lei:2014,barber:2021},
and conditional validity is not implied by A0; the result is therefore asymptotic and transfers
conditional validity from the base sets to the CSL, rather than establishing it. For
$P_{\bX}$-almost every $\bx\in\mathcal X$, let
$P_{n,k}(\bx):=\Pr\big(Y\notin\mathcal C_k(\bx;\mathcal D_n)\mid\mathcal D_n,\bX=\bx\big)$, and
consider the conditional counterparts of marginal validity and of A1$^\star$:
\begin{itemize}
\item[A1$(\bx)$.] For each $k\in\{1,\dots,K\}$,
$\limsup_{n\to\infty}\Pr\big(Y\notin\mathcal C_k(\bX;\mathcal D_n)\mid\bX=\bx\big)\le\alpha$.
\item[A1$^\star(\bx)$.] For each $k\in\{1,\dots,K\}$ and every $\varepsilon>0$,
$\Pr\big(P_{n,k}(\bx)>\alpha+\varepsilon\big)\to0$ as $n\to\infty$.
\end{itemize}

\begin{theorem}[Asymptotic conditional validity]\label{th:conditionalvalidity}
Let the test pair $(\bX,Y)$ be independent of $\mathcal D_n$, and consider the Split-CSL or
Full-CSL under A3. Then, for almost every $\bx\in\mathcal X$:
\begin{enumerate}
\item[(a)] \textbf{General case.} Under A1$^\star(\bx)$,
\[
\liminf_{n\to\infty}\Pr\big(Y\in\mathcal C_{\mathrm{comb}}(\bX;\mathcal D_n)\mid\bX=\bx\big)\ge1-2\alpha .
\]
\item[(b)] \textbf{Dominant learner.} Under A2, if there exists $k^\star$ such that
$\inf_{\mathbf w\in\mathcal W^*}w_{k^\star}>1/2$ and A1$(\bx)$ holds for $k=k^\star$, then
\[
\liminf_{n\to\infty}\Pr\big(Y\in\mathcal C_{\mathrm{comb}}(\bX;\mathcal D_n)\mid\bX=\bx\big)\ge1-\alpha .
\]
\item[(c)] \textbf{Coalition.} Assume A2, and suppose there exist a fixed subset
$\mathcal K^*\subseteq\{1,\dots,K\}$ and $\delta>0$ such that A1$(\bx)$ holds for the members of
$\mathcal K^*$,
\[
\inf_{\mathbf w\in\mathcal W^*}\sum_{k\in\mathcal K^*}w_k\ge\tfrac12+\delta
\qquad\text{and}\qquad
\Pr\big(A^C_{n,j}\,\triangle\,A^C_{n,k}\mid\bX=\bx\big)\to0\quad\text{for all } j,k\in\mathcal K^*.
\]
Then
\[
\liminf_{n\to\infty}\Pr\big(Y\in\mathcal C_{\mathrm{comb}}(\bX;\mathcal D_n)\mid\bX=\bx\big)\ge1-\alpha .
\]
\end{enumerate}
\end{theorem}


Marginal validity of the base sets does not imply A1$(\bx)$ or A1$^\star(\bx)$: for instance,
split conformal sets with absolute-residual scores are marginally valid but over- or under-cover
at particular covariate values under heteroscedasticity. The conditions hold, under regularity
conditions, for learners that consistently estimate the conditional distribution of $Y$ given
$\bX$, combined with distributional or quantile-residual non-conformity scores
\citep{chernozhukov:2021,wu:2025}. Part~(a) therefore applies to libraries in which every learner
has this property, whereas parts~(b) and~(c) require it only for the dominant learner or the
members of the coalition.
Finally, for any weights, the CSL set contains the intersection and is contained in the union of
the learner-specific sets, so its width lies between theirs.
\section{Simulation Study}\label{sec:simulation}

In this section, we evaluate the finite-sample behavior, empirical marginal coverage at the 90\% nominal level ($\alpha = 0.1$) and the average width, of the prediction intervals based on the CSL method through a comprehensive simulation across four data-generating processes (DGPs). The experiment is designed as it follows.

\paragraph{Response generation:} Each DGP represents a distinct regression scenario that challenges different aspects of model specification and prediction:

\begin{enumerate}
    \item \textbf{Homoscedastic linear model (S1):} The covariate vectors $\bx_i = (x_{i1},x_{i2},x_{i3})^{\top}$, $i=1,\dots,n$, are generated from a trivariate normal distribution: $\bx_i \stackrel{i.i.d.}{\sim} N_3 (\textbf{0},\Sigma)$, where $\Sigma$ is a compound symmetric matrix with $\Sigma_{jk} = 1$ if $j=k$, and $\Sigma_{jk} = 0.5$ if $j \neq k$. Let $\bX=(X_1,X_2,X_3)^\top \sim N_3 (\textbf{0},\Sigma)$ and $\epsilon \sim  N(0,0.75^2)$ denote generic random variables, with $(\bx_i,\epsilon_i)$ their i.i.d.\ realizations.

    \begin{equation*}
        y_i = 1 + 0.5 x_{i1} - 0.4 x_{i2} + 0.6 x_{i3} + \epsilon_i,
        \quad \epsilon_i \stackrel{i.i.d.}{\sim} {N}(0, 0.75^{2}).
    \end{equation*}

    In this scenario, the variance of the conditional mean $\mu(\bX)$ satisfies $\text{Var}\{\mu(\bX)\}=\text{Var}\{1+0.5X_1 - 0.4X_2 + 0.6X_3\}=0.63$, while the noise variance is $\text{Var}(\epsilon)=0.75^2$, producing a signal-to-noise ratio $\text{SNR}=1.12$. This places the model in a non-degenerate signal-to-noise regime, avoiding extreme leverage effects and ensuring numerical stability across Monte Carlo replications.

    \item \textbf{Homoscedastic non-linear model (S2):} The covariates are generated as in S1, but the conditional mean is modified to include a non-linear component in $x_{i3}$:

    \begin{equation*}
        y_i = 1 + 0.5 x_{i1} - 0.4 x_{i2} + 0.5 \times (0.6 x_{i3}^{3}) + \epsilon_i,
        \quad \epsilon_i \stackrel{i.i.d.}{\sim} {N}(0, 0.75^{2}).
    \end{equation*}

    In this scenario, the variance of the conditional mean is $\text{Var}\{\mu(\bX)\} = \text{Var}\{1+0.5X_1 - 0.4X_2 + 0.5 \times (0.6X_3^3)\} =1.65$, while the noise variance remains $\text{Var}(\epsilon)=0.75^2$, resulting in $\text{SNR}\approx2.93$. The scaling of the non-linear component ensures that the signal strength remains non-degenerate and comparable to S1, while introducing smooth misspecification of the mean structure without inducing excessive variability.

    \item \textbf{Heteroscedastic linear model (S3):} The continuous covariates $x_{i1}$, $x_{i2}$, and $x_{i3}$ are generated as in S1, together with an additional independent binary covariate $x_{i4} \sim Bernoulli(0.5)$. The conditional mean is linear:

    \begin{equation*}
        \mu(x_{i1}, x_{i2}, x_{i3}) = 3(1 + 0.5 x_{i1} - 0.4 x_{i2} + 0.6 x_{i3}),
    \end{equation*}

    and the conditional standard deviation depends on all covariates:

    \begin{equation*}
        \sigma(x_{i1}, x_{i2}, x_{i3}, x_{i4}) = \text{exp}(\text{log}(0.75) + 0.25x_{i1} + 0.08x_{i2} + 0.18x_{i3} + 0.9x_{i4}).
    \end{equation*}

    The baseline variability equals 0.75, matching the homoscedastic scenarios. The response is generated as:

    \begin{equation*}
        y_i = \mu(x_{i1}, x_{i2}, x_{i3}) + \sigma(x_{i1}, x_{i2}, x_{i3}, x_{i4})\epsilon_i, \quad \epsilon_i \stackrel{i.i.d.}{\sim} {N}(0, 1).
    \end{equation*}
    In this scenario, the variance of the conditional mean satisfies $\text{Var}\{\mu(\bX)\}=9\,\text{Var}\{1+0.5X_1-0.4X_2+0.6X_3\}=5.67$. Since the noise variance is covariate-dependent, we characterize signal strength through the marginal $\text{SNR}=\text{Var}\{\mu(\bX)\}/\mathbb E\{\sigma^2(\bX,X_4)\}$, which equals approximately $1.99$ under the specified design. This places the model in a non-degenerate signal-to-noise regime while inducing substantial heteroscedasticity through both continuous covariates and the binary grouping variable.

    \item \textbf{Sparse linear model with spurious covariates (S4):} The response follows the homoscedastic linear model of S1, depending only on $(x_{i1}, x_{i2}, x_{i3})$:

    \begin{equation*}
        y_i = 1 + 0.5 x_{i1} - 0.4 x_{i2} + 0.6 x_{i3} + \epsilon_i,
        \quad \epsilon_i \stackrel{i.i.d.}{\sim} {N}(0, 0.75^{2}).
    \end{equation*}

    In addition to these active covariates, ten spurious variables are generated independently of both the response and the active predictors: five continuous variables from $N(0,1)$ and five binary variables from $\text{Bernoulli}(0.5)$. All covariates are provided as inputs to each candidate learner, although only three enter the true data-generating mechanism. This scenario is designed to assess the robustness of the proposed method to irrelevant covariates and high-dimensional noise. Since the data-generating mechanism coincides with S1, and $\text{SNR}=1.12$.

\end{enumerate}

\paragraph{Sample sizes and replications.} Sample sizes considered are $n \in \{100, 500, 1000\}$. For each simulation setting, the procedure is repeated across 1,000 Monte Carlo replications to evaluate empirical coverage. In each replication, a new test point is generated to assess whether the constructed prediction interval contains the true response value and the interval width.

\paragraph{CSL construction:} The SL library includes five statistical models or algorithms: 

        \begin{itemize}
        
            \item \textbf{Linear model (LM):} Ordinary least squares regression assuming a linear conditional mean and homoscedastic Gaussian errors, implemented with the \texttt{stats} package in R.
            
            \item \textbf{Generalized additive model (GAM):} A semiparametric regression model with the conditional mean expressed as an additive sum of smooth functions of covariates, fitted via penalized splines and implemented with the \texttt{mgcv} package in R.
        
            \item \textbf{Generalized additive model for location, scale and shape (GAMLSS):} A distributional regression model that specifies parametric conditional distributions. Here, a normal distribution is assumed with additive predictors for both mean and variance, explicitly accounting for heteroscedasticity. Implemented with the \texttt{gamlss} package in R.
            
            \item \textbf{Neural network (NNET):} A single-hidden-layer feedforward neural network used as a flexible non-linear estimator of the conditional mean, implemented with the \texttt{nnet} package in R.
            
            \item \textbf{Random forest (RF):} An ensemble of regression trees constructed via bootstrap aggregation and random feature selection, providing a nonparametric estimator of the conditional mean, implemented with the \texttt{randomForest} package in R.

            For scenario S4 (sparse linear model with spurious covariates), we replace GAMLSS with a regularized linear model to better handle high-dimensional noise:
    
            \begin{itemize}
                \item \textbf{Least Absolute Shrinkage and Selection Operator (LASSO):} Linear regression with $\ell_1$ penalty, implemented with the \texttt{glmnet} package in R using ten-fold cross-validation to select the regularization parameter.
            \end{itemize}
        
        \end{itemize}

Accordingly, the SL library comprises LM, GAM, NNET, RF, and GAMLSS in scenarios S1–S3, and LM, GAM, NNET, RF, and LASSO in scenario S4. The dimensionality of the covariate space varies across scenarios: three covariates are used in S1 and S2, four in S3, and thirteen in S4, and all candidate learners are fitted using the same covariate set within each scenario.
In the Split-CSL implementation, the sample is evenly partitioned into training and calibration sets. This 50:50 split serves as a benchmark against the Full-CSL procedure: only half the data are used to fit the base learners and estimate the SL weights, whereas the full conformal method uses the entire sample. Consequently, differences between the two approaches are more visible under this design.
Treating these as candidate learners and following the process introduced in Section~\ref{subsec:SL}, the SL weights are obtained via 5-fold cross-validation and non-negative least squares, implemented with the \texttt{nnls} package in R. For each candidate, we then construct both split and full conformal prediction intervals, where the interval-finding techniques are similar to \cite{wu:2025}. For the GAMLSS learner, non-conformity scores are based on quantile residuals to account for heteroscedasticity \citep{wu:2025}; for the remaining learners (including LASSO in S4), absolute raw residuals are used. Finally, these individual conformal prediction intervals are combined using the weighted majority vote rule defined in Section~\ref{subsec:CP_aggregation}, producing the CSL prediction intervals.

\subsection*{Simulation results}

Table \ref{tab:coverage} summarizes empirical marginal coverage probabilities and average interval widths for the ``Correct Model'' (true DGP) and CSL procedures under split and full conformal prediction across the four data-generating scenarios and sample sizes. The correct model benchmark corresponds to conformal prediction intervals constructed using the correct model for each scenario, that is, the model corresponding to the true data-generating mechanism. 
Overall, the CSL achieves empirical coverage close to the nominal 90\% level across all scenarios and sample sizes under both split and full conformal settings. This performance notably exceeds the conservative worst-case lower bound of $1-2\alpha$ (80\% for $\alpha=0.1$) established for weighted majority vote aggregation in Theorem \ref{th:splitvalidity}(a). In terms of efficiency, the CSL produces interval widths comparable to those of the correct model benchmark. As expected, widths generally decrease with increasing sample size $n$, and scenario S3 leads to wider intervals due to its covariate-dependent noise structure, with the CSL remaining competitive relative to the correct model in all cases.
To examine the weight allocation behavior of the CSL, Table \ref{tab:weights} reports, for each scenario, the percentages of simulations where the most frequently dominant and preferred candidate achieved weight $>0.5$ (denoted by Dominant \%) and attained the highest weight (denoted by Preferred \%). The candidate identified in parentheses is the one that most frequently achieved these statuses across simulations for that scenario.

\begin{table}[!htbp]
\small
\centering
\begin{tabular}{llccccc}
\toprule
\multirow{2}{*}{\textbf{Scenario}} & \multirow{2}{*}{\textbf{Sample Size}} & \multicolumn{2}{c}{\textbf{Split Conformal}} & \multicolumn{2}{c}{\textbf{Full Conformal}} \\
\cmidrule(lr){3-4} \cmidrule(lr){5-6}
 & & \textbf{Correct Model} & \textbf{CSL} & \textbf{Correct Model} & \textbf{CSL} \\
\midrule
\multirow{3}{*}{\textbf{S1}} 
 & n = 100  & 0.913/2.6459 & 0.914/2.7811 & 0.909/2.5594 & 0.911/2.5868 \\
 & n = 500  & 0.912/2.4957 & 0.913/2.5119 & 0.909/2.4821 & 0.907/2.4821 \\
 & n = 1000 & 0.921/2.4839 & 0.916/2.4916 & 0.913/2.4749 & 0.911/2.4766 \\
\midrule
\multirow{3}{*}{\textbf{S2}} 
 & n = 100  & 0.910/2.9955 & 0.913/3.1769 & 0.903/2.9226 & 0.901/2.9512 \\
 & n = 500  & 0.908/2.5588 & 0.908/2.5624 & 0.906/2.5260 & 0.905/2.5280 \\
 & n = 1000 & 0.917/2.5192 & 0.915/2.5227 & 0.910/2.4966 & 0.911/2.4933 \\
\midrule
\multirow{3}{*}{\textbf{S3}} 
 & n = 100  & 0.929/6.1990 & 0.914/6.4916 & 0.914/5.0930 & 0.901/5.1742 \\
 & n = 500  & 0.903/4.8944 & 0.902/4.9942 & 0.900/4.7837 & 0.904/4.8745 \\
 & n = 1000 & 0.913/4.8094 & 0.913/4.9186 & 0.910/4.7669 & 0.904/4.8818 \\
\midrule
\multirow{3}{*}{\textbf{S4}} 
 & n = 100  & 0.913/2.6459 & 0.910/3.0015 & 0.909/2.5594 & 0.914/2.6817 \\
 & n = 500  & 0.912/2.4957 & 0.907/2.5433 & 0.909/2.4821 & 0.910/2.5004 \\
 & n = 1000 & 0.921/2.4839 & 0.912/2.5036 & 0.913/2.4749 & 0.909/2.4849 \\
\bottomrule
\end{tabular}
\caption{\textbf{Empirical coverage/average interval width for the correct model and CSL under split and full conformal prediction.} The correct model for each scenario is the model aligned with the true data-generating mechanism (LM for S1, GAM for S2, GAMLSS for S3, and LM for S4).}
\label{tab:coverage}
\end{table}

\begin{table}[!htbp]
\small
\centering
\begin{tabular}{llccccc}
\toprule
\multirow{2}{*}{\textbf{Scenario}} & \multirow{2}{*}{\textbf{Sample Size}} 
& \multicolumn{2}{c}{\textbf{Split Conformal}} 
& \multicolumn{2}{c}{\textbf{Full Conformal}} \\
\cmidrule(lr){3-4} \cmidrule(lr){5-6}
 & & Dominant (\%) & Preferred (\%) & Dominant (\%) & Preferred (\%) \\
\midrule
\multirow{3}{*}{\textbf{S1 (LM)}} 
 & n = 100  & 49.2 & 53.9 & 49.2 & 52.0 \\
 & n = 500  & 47.3 & 49.3 & 44.4 & 45.4 \\
 & n = 1000 & 47.5 & 49.4 & 43.8 & 45.1 \\
\midrule
\multirow{3}{*}{\textbf{S2 (GAM)}} 
 & n = 100  & 55.8 & 60.4 & 92.0 & 94.3 \\
 & n = 500  & 97.8 & 98.0 & 95.3 & 95.6 \\
 & n = 1000 & 95.8 & 95.9 & 91.7 & 92.0 \\
\midrule
\multirow{3}{*}{\textbf{S3 (GAMLSS)}} 
 & n = 100  & 72.3 & 77.7 & 82.1 & 86.4 \\
 & n = 500  & 85.7 & 88.4 & 85.0 & 88.1 \\
 & n = 1000 & 85.3 & 89.5 & 86.3 & 88.4 \\
\midrule
\multirow{3}{*}{\textbf{S4 (LASSO)}} 
 & n = 100  & 38.5 & 44.8 & 58.3 & 64.5 \\
 & n = 500  & 73.3 & 76.8 & 77.5 & 80.5 \\
 & n = 1000 & 80.6 & 83.3 & 79.8 & 83.2 \\
\bottomrule
\end{tabular}
\caption{\textbf{Weight allocation behavior of CSL: dominant and preferred percentages for the most frequently selected candidate in each scenario.}}
\label{tab:weights}
\end{table}

The weight allocation patterns provide insight into the empirical performance of the CSL. In scenarios S2, S3, and S4, both dominant and preferred percentages increase with sample size, indicating that the CSL progressively concentrates weight on a single candidate. By $n=1000$, the most frequently selected candidate receives a dominant weight (exceeding $0.5$) in approximately $80$--$96\%$ of simulations. This behavior is broadly consistent with the sufficient condition in Theorem~\ref{th:marginalvalidity}(b), under which the ensemble attains marginal coverage of at least $1-\alpha$ when a single candidate asymptotically dominates, a pattern that becomes more evident as the sample size increases. 
Scenario~S1 exhibits a distinct pattern. Although LM remains the most frequently dominant and preferred candidate, its dominant frequency under full conformal prediction decreases modestly with sample size, from $49.2\%$ at $n=100$ to $43.8\%$ at $n=1000$. This decline is accompanied by a corresponding increase in the dominance of the GAMLSS candidate, whose dominant percentage rises from $32.4\%$ to $41.7\%$. This behavior reflects the flexibility of the GAMLSS model, which contains the linear Gaussian model as a special case and can achieve predictive performance comparable to that of LM once sufficient data are available. Consequently, the CSL distributes weight across multiple competitive candidates rather than concentrating on a single learner. Importantly, this redistribution does not compromise coverage or efficiency. Even in Scenario~S1, where no candidate appears to dominate consistently, the CSL maintains empirical coverage close to the nominal level and produces prediction intervals only marginally wider than those of the correct model (Table~\ref{tab:coverage}). This pattern seems to be consistent with Theorem~\ref{th:marginalvalidity}(c), which permits asymptotic marginal validity at level $1-\alpha$ when a coalition of competitive learners, rather than a single dominant learner, jointly receives sufficient weight and collectively determines the weighted majority vote. Under both split and full conformal settings, the combined coalition (LM and GAMLSS) weight exceeds 0.5 over 80\% of replications across all sample sizes considered.
Overall, the simulation study demonstrates that the CSL delivers valid and efficient prediction intervals in finite samples. It adapts its weight allocation to the underlying data-generating structure, frequently satisfying the conditions for the tighter $1-\alpha$ coverage guarantee, while remaining robust when multiple candidates exhibit comparable predictive performance.

\section{Case study: NHANES data}\label{sec:application}

We analyze a data set from the National Health and Nutrition Examination Survey (NHANES) August 2021--August 2023 cycle \cite{nhanes20212023}. NHANES provides nationally representative sociodemographic, clinical, and nutritional data on U.S. adults and children, making it a widely used resource for monitoring population health and for predictive modeling \citep{matabuena:2025}. 
We take serum creatinine, \texttt{LBXSCR} (mg/dL), as the response variable, as it is a key biomarker of kidney function and a central component in the estimation of glomerular filtration rate (eGFR). Accurate prediction of creatinine levels is therefore important for the early detection and monitoring of kidney function. 
We exclude participants with kidney conditions, as impaired renal function directly alters creatinine levels. We similarly exclude pregnant participants, whose creatinine levels are systematically lower \citep{davison:1974}.
We include 18 covariates in total, comprising 12 continuous and 6 categorical variables, as described in the Appendix. The selected covariates are motivated by their established associations with serum creatinine through demographic, anthropometric, and clinical pathways. After selecting the response and covariates, observations with missing values in any of these variables are removed, resulting in a final data set of $n = 5027$ observations. The response ranges from 0.35 to 4.48, with a mean of 0.89. A summary of the covariates is presented in the Appendix. Since creatinine levels can only take positive values, all candidate learners are fitted to the log-transformed creatinine values. The resulting conformal prediction intervals are subsequently back-transformed to the original scale via exponentiation to ensure interpretability.

We consider a SL library consisting of six candidate models: LM, LASSO, GAM, NNET, RF, and GAMLSS, introduced in Section \ref{sec:simulation}, spanning modeling approaches from simple linear regression to flexible non-linear and distributional (heteroscedastic) methods. For the split conformal setting, the data set is randomly partitioned into training, calibration, and testing sets, with 10\% of observations reserved for testing and the remaining 90\% divided between the training and calibration sets in an 80:20 ratio. The Appendix presents a sensitivity analysis to other splitting rules, showing that they all lead to empirical coverage above the nominal \(90\%\) level, but exhibiting some variability in terms of the estimated SL weights. Table \ref{tab:sl_weights} reports the estimated SL weights under the split and full conformal settings, together with the empirical coverage and mean width of the corresponding 90\% conformal prediction intervals on the testing set.
Both conformal procedures achieve coverage at or above the nominal \(90\%\) level, with the full conformal method producing narrower intervals than the split approach. As a benchmark, classical prediction intervals based on ordinary least squares linear regression fitted to the combined training and calibration data attain empirical coverage of \(0.916\) and mean width \(0.5556\). Although these intervals also achieve the desired coverage, they are substantially wider, indicating that the conformalized SL methods provide more efficient prediction intervals while maintaining the target coverage level.

RF receives more than half of the weight in both settings (Table~\ref{tab:sl_weights}), so the CSL set coincides with the conformal prediction set of RF, while GAM and LM share most of the
remaining weight. The cross-validated predictions of LM, GAM, LASSO, GAMLSS and RF are highly correlated (Pearson correlations above $0.93$), and RF attains the smallest cross-validated RMSE
($0.1907$, against $0.1923$ for GAM and $0.1930$ for LM). Under non-negative least squares, such strongly correlated predictions are largely redundant, so the weight concentrates on RF, GAM and
LM, while NNET, with the largest RMSE ($0.2204$), receives none. A plausible explanation for the advantage of RF is its ability to capture non-linear effects and interactions among demographic,
anthropometric and laboratory variables without explicit specification.

\begin{figure}[!htbp]
  \centering
  \includegraphics[width=0.75\textwidth]{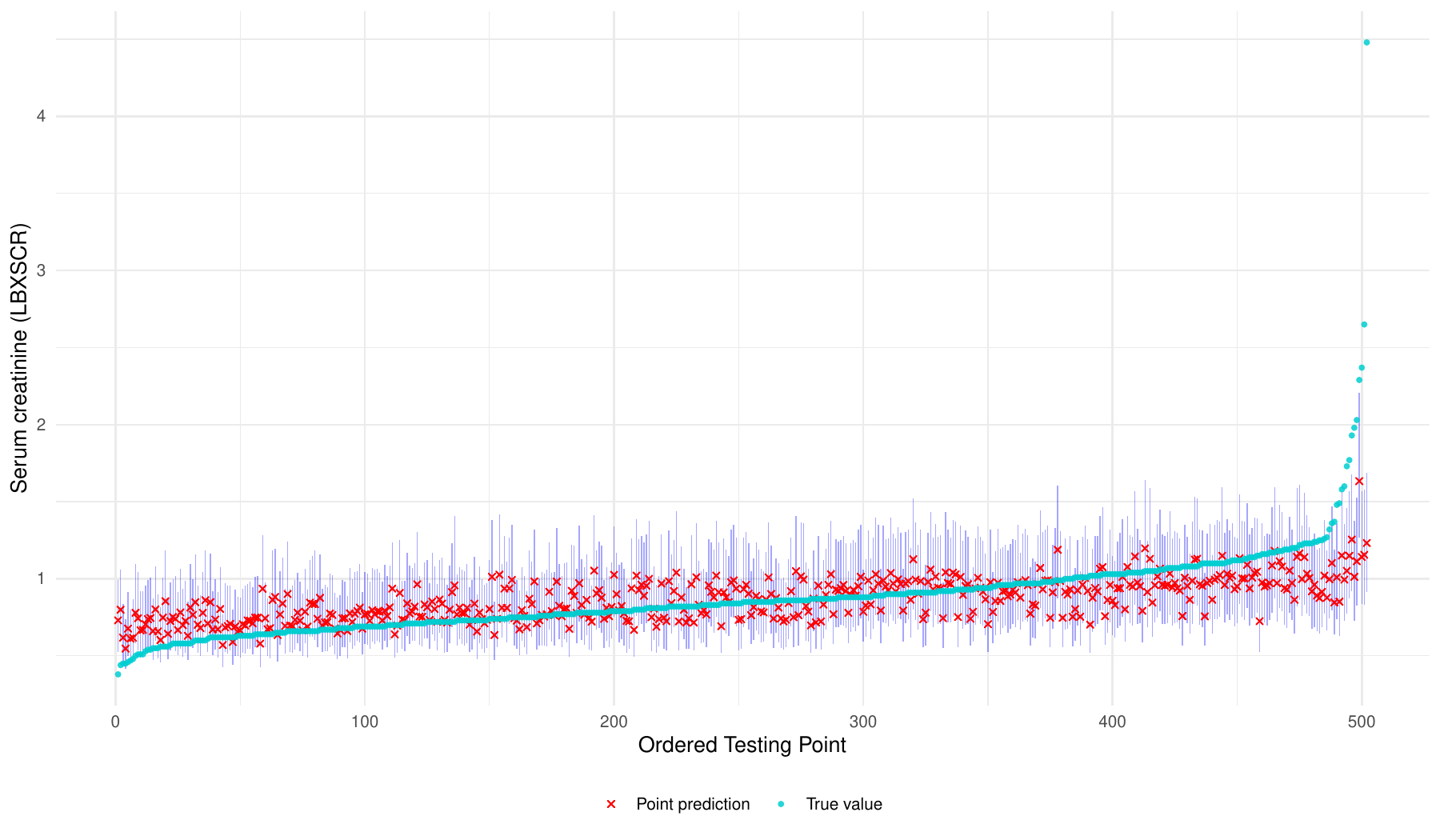}
  \caption{\textbf{Full-CSL prediction intervals for testing observations ordered by the observed response.}}
  \label{fig:testing_performance}
\end{figure}

For brevity, the main text focuses on the Full-CSL results; the corresponding Split-CSL plots are reported in the Appendix. Figure \ref{fig:testing_performance} displays the Full-CSL prediction intervals for the testing observations, ordered by observed serum creatinine values. Overall, the intervals track the increasing trend in the response reasonably well, and the point predictions follow the central location of the observed values across most of the sample. Interval widths remain fairly stable over much of the ordered sequence but become more variable toward the upper tail, reflecting greater predictive uncertainty for larger creatinine values. The figure also shows that several of the most extreme observations lie substantially above both the point predictions and the corresponding upper interval bounds. This suggests the presence of unusual response values not adequately explained by the current covariates, indicating that additional unobserved or omitted factors may contribute to variation in the upper tail.
The Appendix presents covariate-specific Full-CSL prediction interval curves for serum creatinine based on three representative testing profiles. Among the covariates, uric acid (\texttt{LBXSUA}) shows the strongest and most consistent positive association: across all three profiles, both the point prediction and the interval bounds increase clearly as uric acid rises. Age (\texttt{RIDAGEYR}) also exhibits a positive effect, particularly for the high and low observed profiles, while the increase is more modest for the middle profile. Most of the remaining covariates generate nearly horizontal curves, although a few show small profile-specific fluctuations. Overall, the Full-CSL curves suggest that the main marginal contributors to predicted serum creatinine are uric acid and age, while the effects of the other covariates are comparatively limited when the remaining variables are held fixed.


\subsection{Sensitivity to extreme observations}

Visual inspection of the response distribution reveals marked right-skewness, with a small number of observations lying far above the bulk of the data; the corresponding boxplot is provided in the Appendix (Figure \ref{fig:response_plots}). Serum creatinine values above 1.5 mg/dL are commonly regarded as elevated and may be suggestive of impaired renal function, although clinical interpretation depends on patient-specific factors such as age, sex, and muscle mass. We therefore exclude observations with serum creatinine exceeding 1.5 mg/dL in the sensitivity analysis, as these values may correspond to individuals with compromised kidney function whose creatinine levels are influenced by pathological mechanisms distinct from those governing the broader study population. In total, 104 observations are removed, resulting in a final sample size of 4923.
We repeat the analysis from the previous section after excluding these extreme response values. This outlier-removed analysis is presented solely as a sensitivity analysis to assess the stability and generalizability of the CSL findings for the main analysis population. We do not interpret the excluded observations as erroneous or clinically irrelevant; rather, they may correspond to genuinely severe physiological states.

Table \ref{tab:sl_weights} reports the estimated SL weights under both conformal settings after removing outliers. In both settings, RF receives the largest weight and exceeds the 0.5 threshold, so the aggregation proceeds via the dominant learner in each case. The weights are similar across the two settings: RF receives dominant weights, followed by GAM (around 0.25--0.37) and LM (around 0.07--0.16), while LASSO, NNET, and GAMLSS receive negligible weights. Both settings achieve empirical coverage of the nominal 90\% level, and the average interval widths are smaller than in the main analysis (Table \ref{tab:sl_weights}), reflecting the reduced variability in the response after excluding extreme creatinine values.

\begin{figure}[!htbp]
  \centering
  \includegraphics[width=0.9\textwidth]{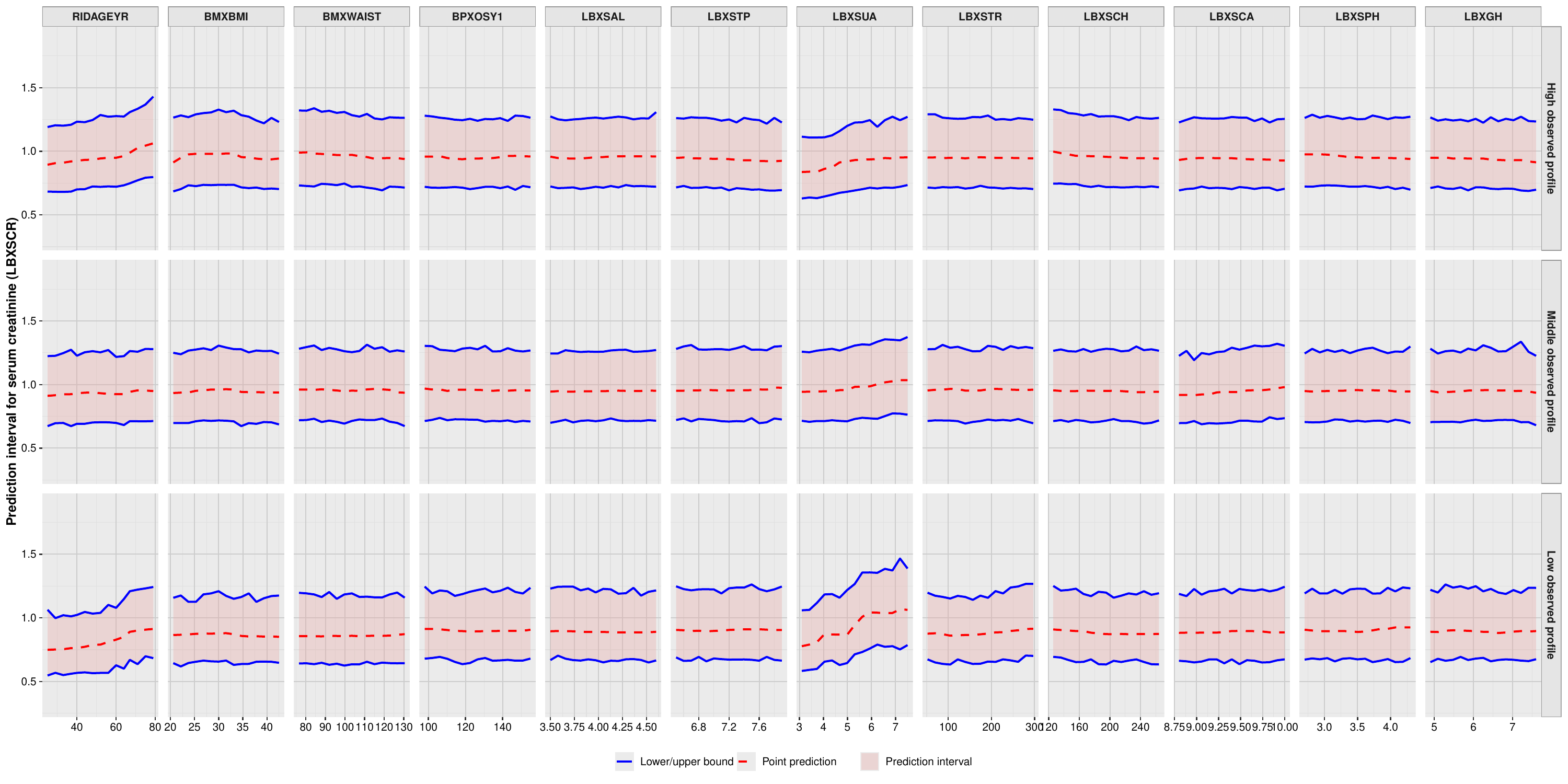}
  \caption{\textbf{Covariate-specific Full-CSL prediction intervals for three representative profiles after removing outliers.}}
  \label{fig:prediction_intervals_covariate_removal}
\end{figure}

Figures \ref{fig:testing_performance_removal} and \ref{fig:split_testing_performance_removal} (in the Appendix) show the prediction intervals for the testing observations after removing outliers under the full and split CP settings, respectively. In both cases, the intervals are more stable across the ordered sequence than in the main analysis, and the discrepancy between the point predictions and observed responses in the upper tail is smaller, as expected. Figures \ref{fig:prediction_intervals_covariate_removal} and \ref{fig:split_prediction_intervals_covariate_removal} (in the Appendix) display the covariate-specific prediction interval curves after outlier removal. Compared with the main analysis, the curves are flatter under both settings, reflecting the reduced range of variation to be explained by the covariates after excluding high creatinine values. The positive trends for uric acid and age remain visible but are weaker (except for the high profile of uric acid), and the interval widths are narrower across the covariate space. Overall, removing the extreme observations reduces instability in the upper tail and narrows the intervals without altering the main conclusions. This also suggests that RF is favored for reasons beyond robustness to outliers, likely due to its ability to capture non-linearities and interaction effects.

\begin{table}[!htbp]
\small
\centering
\begin{tabular}{lccccccc}
\toprule
\textbf{Setting} & \textbf{LM} & \textbf{LASSO} & \textbf{GAM} & \textbf{NNET} & \textbf{RF} & \textbf{GAMLSS} & \textbf{Coverage/Width} \\
\midrule
\textbf{Split conformal} (F) & 0.1457 & 0.0000 & 0.2720 & 0.0000 & 0.5823 & 0.0000 & 0.912/0.5283 \\
\textbf{Full conformal} (F) & 0.1951 & 0.0000 & 0.2336 & 0.0000 & 0.5713 & 0.0000 & 0.914/0.5279 \\
\textbf{Split conformal} (RO) & 0.0697 & 0.0000 & 0.3695 & 0.0000 & 0.5608 & 0.0000 & 0.902/0.4944 \\
\textbf{Full conformal} (RO) & 0.1570 & 0.0000 & 0.2471 & 0.0000 & 0.5959 & 0.0000 & 0.907/0.4883 \\
\bottomrule
\end{tabular}
\caption{\textbf{Estimated SL weights under the split and full conformal settings. (F) denotes the analysis based on the full data set, while (RO) denotes the analysis after removing outliers.}}
\label{tab:sl_weights}
\end{table}

\section{Discussion}\label{sec:discussion}

The proposed Conformalized Super Learner (CSL) equips the Super Learner (SL) with prediction sets by conformalizing each learner separately, with its own non-conformity score, and aggregating the resulting sets through a weighted majority vote with the SL weights. For the Split-CSL, coverage is at least $1-2\alpha$ in finite samples regardless of the weight values, and at least $1-\alpha$ whenever a single learner holds more than half of the weight. For the Full-CSL, analogous guarantees hold asymptotically.
The worst-case bound $1-2\alpha$ arises only in special configurations. For instance, with two learners of weight $1/2$ each, the combined set is the intersection of their sets; if the two sets
fail to cover $Y$ on disjoint events, each with probability $\alpha$, the combined set fails with probability $2\alpha$. Since large errors tend to affect all learners at once, coverage is typically much closer to $1-\alpha$.
Compared with methods that minimize the size of the prediction set \citep{hegazy:2026,xu:2026,liang:2026}, the CSL trades optimal width for interpretability. For both
algorithms, the weights do not depend on the test point and describe how much each learner
contributes, the same weights define the SL point prediction, and the vote shows where the learners
agree; when a single learner holds more than half of the weight, the CSL set is exactly that
learner's set. For the Split-CSL, this dominance is known before the test point is observed, so the
applicable guarantee can be reported with the sets; for the Full-CSL, it is only asymptotic and
requires dominance at the population level. The price is a weaker worst-case guarantee than
selection-corrected methods, which attain $1-\alpha$ in finite samples for any collection of
pretrained models \citep{liang:2026}.

The simulations deliver two messages. First, performance is driven by the library: when it contains
the correct model, or a learner that captures its key features such as non-linearity or
heteroscedasticity, the CSL produces intervals comparable in width to the correct-model benchmark,
and its coverage remained close to the nominal level also in Scenarios S3 and S4, where all learners
included uninformative covariates. Second, both regimes of the theory appear in practice: a single
learner dominates in Scenarios S2--S4, so that the CSL reduces to that learner, whereas in
Scenario~S1 the weight is shared between LM and GAMLSS, which estimate the same linear mean,
consistent with the coalition regime.

The case study illustrates how the proposed CSL can be applied in practice to predict creatinine levels while accounting for a wide range of features within a library of candidate learners. More specifically, the NHANES database contains a large number of covariates of diverse types, including clinical, sociodemographic, and laboratory measurements, which naturally poses challenges for modeling. One may question whether all variables meaningfully contribute to explaining creatinine levels, raising concerns about sparsity. Additionally, the functional role of each covariate, such as potential nonlinear effects and complex interactions, must be considered. These challenges are difficult to capture with a single learner; therefore, it is desirable to incorporate multiple learners capable of modeling different aspects of the data. The proposed CSL methodology thus plays a crucial role in quantifying the uncertainty of ensemble predictions, enabling users to assess both the variability of these predictions and the importance of the available variables in explaining the outcome of interest. Among the 18 variables considered, we find that only a subset exhibits strong predictive power. This insight may be valuable for clinicians and epidemiologists in identifying the primary drivers of elevated creatinine levels associated with kidney disease. Beyond valid coverage, the Full-CSL achieves comparable coverage to ordinary least squares (91.4\% vs. 91.6\%) while reducing mean interval width by approximately 5\% (0.528 vs. 0.556), demonstrating meaningful efficiency gains from the ensemble approach. This application also presents a case in which one learner dominates the ensemble, with a weight exceeding $0.5$. This allows for further inspection via the corresponding dominant learner (RF) and its associated diagnostic tools. 

Our work opens several avenues for future research. From a methodological perspective, one could consider broader classes of ensemble methods, as well as alternative aggregation strategies beyond the weighted majority vote considered here. While we have focused on continuous outcomes, natural extensions include discrete outcomes (\textit{e.g.}, binary and Poisson regression), bounded outcomes \citep{wu:2025}, and survival outcomes \citep{golmakani:2020,li:2025,hothorn:2006}. In this work, we have focused on the use of \emph{tailor-made} non-conformity scores, that is, those that take the relevant features of the candidate learner into account, such as heteroscedasticity. A potential extension to automate this choice is to use conformal quantile regression (CQR) \citep{romano:2019,jensen:2022,susmann:2025} for each candidate learner. This approach is robust across different types of learners, but it is not necessarily tailored to any specific one. A natural extension to heterogeneous model libraries would be to employ learner-specific non-conformity scores for interpretable models (\textit{e.g.}, residual-based scores for distributional regression models \citep{chernozhukov:2018}) and CQR for \emph{black-box} learners (such as random forests or neural networks). A systematic comparison with size-minimizing and
selection-corrected methods, including coverage within subgroups, would clarify when
interpretability is worth its worst-case price.

\bibliography{references}

\clearpage
\section*{Appendix}

\subsection*{Algorithsm}

\begin{algorithm}[!htbp]
\spacingset{1}
\caption{Split Conformalized Super Learning (Split-CSL)}
\label{alg:splitCSL}
\begin{algorithmic}[1]
\small

\Require data set $\left(y_i, \bx_i\right)^n_{i=1}$, new covariate $\bx_{\text{new}}$, miscoverage level $\alpha$, base learners $\mathcal{L}_1, \dots, \mathcal{L}_K$, number of folds V.
\Ensure Split-CSL prediction interval $\mathcal{C}^{\text{split}}_{\text{comb}}(\bx_{\text{new}})$ for the new response $y_{\text{new}}$ corresponding to $\bx_{\text{new}}$.

\State Randomly split the data set into disjoint subsets $\mathcal{I}_{\text{train}}$ (training set) and $\mathcal{I}_{\text{cal}}$ (calibration set).

\State \textbf{(A) SL weight estimation on $\mathcal{I}_{\text{train}}$}

\State Construct the cross-validated empirical risk $R_n(\bw)$ on $\mathcal{I}_{\text{train}}$ as in Section~\ref{subsec:SL}.
\State Compute $\widehat{\bw}_n = \left( \widehat{w}_{n,1}, \dots, \widehat{w}_{n,K} \right) = \operatorname{argmin}_{\bw \in \Delta^{K-1}} R_n(\bw)$.

\State \textbf{(B) Split conformal base intervals:}

\For{$k = 1,\dots,K$}
    \State Fit learner $\mathcal{L}_k$ on $\{(y_i,\bx_i): i \in \mathcal{I}_{\text{train}}\}$ and obtain predictor $\widehat f_k$.
    \State Compute predictions $\widehat f_k(\bx_i)$ for $i \in \mathcal{I}_{\text{cal}}$ and $\widehat f_k(\bx_{\text{new}})$.
    \State Compute non-conformity scores $s_{i,k} = \text{score}\big(y_i, \widehat f_k(\bx_i)\big)$ for all $i \in \mathcal{I}_{\text{cal}}$
    \State Let $q_{k,\alpha}$ be the $\left\lceil (1-\alpha)(\#\mathcal{I}_{\text{cal}}+1)\right\rceil$-th smallest value of $\{ s_{i,k} : i \in \mathcal{I}_{\text{cal}} \}$, where $\#\cdot$ denotes the set size.
    \State Define the prediction interval as $\mathcal{C}_k(\bx_{\text{new}}) = \left[ L_k(\bx_{\text{new}}), U_k(\bx_{\text{new}}) \right] =\{y \in \mathbb{R} : \text{score}\left(y, \widehat f_k(\bx_{\text{new}})\right)\le q_{k,\alpha}\}$.
\EndFor

\State \textbf{(C) Weighted-majority-vote aggregation:}

\State Define the active set of learners with positive weights: $\mathcal{K}_{\text{act}}=\{k\in\{1,\dots,K\}:\widehat w_{n,k}>0\}$.

\State {\textbf{(C1) A dominant learner exists:}}
\State If there exists $k^\star\in\mathcal{K}_{\text{act}}$ such that $\widehat w_{n,k^\star}>1/2$, set $\mathcal{C}^{\text{split}}_{\text{comb}}(\bx_{\text{new}})
=\mathcal{C}_{k^\star}(\bx_{\text{new}})$, and end the algorithm.

\State {\textbf{(C2) No dominant learner exists:}}
\State Construct a one-dimensional grid $\mathcal{G}$ over the union of interval endpoints from active learners.

\For{each $y_j \in \mathcal{G}$}
\State Compute the weighted vote score $V(y_j)=\sum_{k\in\mathcal{K}_{\text{act}}}\widehat w_{n,k}\,
    \mathbf{1}_{\left\{L_k(\bx_{\text{new}})\le y_j\le U_k(\bx_{\text{new}})\right\}}$.
\EndFor

\State Obtain the Split-CSL prediction set $\mathcal{C}^{\text{split}}_{\text{comb}}(\bx_{\text{new}})=\{y_j\in\mathcal{G}: V(y_j)>1/2\}$.



\end{algorithmic}
\end{algorithm}

\begin{algorithm}[!htbp]
\spacingset{1}
\caption{Full Conformalized Super Learning (Full-CSL)}
\label{alg:fullCSL}
\begin{algorithmic}[1]
\small

\Require data set $\left(y_i, \bx_i\right)^n_{i=1}$, new covariate $\bx_{\text{new}}$, miscoverage level $\alpha$, base learners $\mathcal{L}_1, \dots, \mathcal{L}_K$, number of folds V.
\Ensure Full-CSL prediction interval $\mathcal{C}^{\text{full}}_{\text{comb}}(\bx_{\text{new}})$ for the new response $y_{\text{new}}$ corresponding to $\bx_{\text{new}}$.

    \State \textbf{(A) SL weight estimation}
    
    \State Construct the cross-validated empirical risk $R_n(\bw)$ on the data set $\left( y_i, \bx_i \right)^n_{i=1}$ as in Section~\ref{subsec:SL}.
    \State Compute $\widehat{\bw}_n = \left( \widehat{w}_{n,1}, \dots, \widehat{w}_{n,K} \right) = \operatorname{argmin}_{\bw \in \Delta^{K-1}} R_n(\bw)$.
    
    \State \textbf{(B) Full conformal base intervals}

    \For{$k = 1,\dots,K$}

        \State Define a grid $\mathcal{Y} \subset \mathbb{R}$ of candidate values for $y_{\text{new}}$.
        \State Initialize an empty set $\mathcal{S} = \emptyset$ to store conformal candidates.
        \For{each $y \in \mathcal{Y}$}
        \State Augment the data set $\left( y_i, \bx_i \right)^n_{i=1}$ with $(y, \bx_{\text{new}})$ to form $\{(y_1, \bx_1), \dots, (y_n, \bx_n), (y, \bx_{\text{new}})\}$.
        \State Fit learner $\mathcal{L}_k$ on the augmented data set and obtain predictor $\widehat{f}_k$.
        \State Compute predictions $\widehat f_k(\bx_i)$ for $i = 1, \dots, n$ and $\widehat f_k(\bx_{\text{new}})$.
        \State Compute non-conformity scores $s_{i,k} = \text{score}\big(y_i, \widehat f_k(\bx_i)\big)$ for all $i$ and $s_{\text{new}}=\text{score}(y,\widehat f_k(\bx_{\text{new}}))$.
        \State Let $q_{k,\alpha}$ be the $\left\lceil (1 - \alpha)(n + 1) \right\rceil$-th smallest value among $\{s_{1,k}, \dots, s_{n,k}\}$.
        \If{$s_{\text{new}} \leq q_{k,\alpha}$}
        \State Include $y$ in $\mathcal{S}$.
        
        \EndIf
        \EndFor
    \State Define the prediction interval as $\mathcal{C}_k(\bx_{\text{new}})=\left[L_k(\bx_{\text{new}}),U_k(\bx_{\text{new}})\right]$ to be lower and upper bounds of \( \mathcal{S} \).
    \EndFor

\State \textbf{(C) Weighted-majority-vote aggregation:}

\State Define the active set of learners with positive weights: $\mathcal{K}_{\text{act}}=\{k\in\{1,\dots,K\}:\widehat w_{n,k}>0\}$.

\State {\textbf{(C1) A dominant learner exists:}}
\State If there exists $k^\star\in\mathcal{K}_{\text{act}}$ such that $\widehat w_{n,k^\star}>1/2$, set $\mathcal{C}^{\text{full}}_{\text{comb}}(\bx_{\text{new}})
=\mathcal{C}_{k^\star}(\bx_{\text{new}})$, and end the algorithm.

\State {\textbf{(C2) No dominant learner exists:}}
\State Construct a one-dimensional grid $\mathcal{G}$ over the union of interval endpoints from active learners.

\For{each $y_j \in \mathcal{G}$}
\State Compute the weighted vote score $V(y_j)=\sum_{k\in\mathcal{K}_{\text{act}}}\widehat w_{n,k}\,
    \mathbf{1}_{\left\{L_k(\bx_{\text{new}})\le y_j\le U_k(\bx_{\text{new}})\right\}}$.
\EndFor

\State Obtain the Full-CSL prediction set $\mathcal{C}^{\text{full}}_{\text{comb}}(\bx_{\text{new}})=\{y_j\in\mathcal{G}: V(y_j)>1/2\}$.

\end{algorithmic}
\end{algorithm}

\clearpage

\subsection*{Proof of Theorem \ref{theorem:set_consistency}}


Let $\epsilon > 0$ be arbitrary. We wish to show that $\Pr(d(\widehat{\mathbf{w}}_n, \mathcal{W}^*) \geq \varepsilon) \to 0$ as $n \to \infty$.

By the \textbf{well-separateness} condition (Condition 2), for this $\varepsilon$, there exists an $\eta > 0$ such that for all $\mathbf{v} \in \Delta^{K-1}$:
\begin{equation*}
    d(\mathbf{v}, \mathcal{W}^*) \geq \varepsilon \implies R(\mathbf{v}) \geq R(\mathbf{w}^*) + \eta,
\end{equation*}
where $R(\mathbf{w}^*) = \inf_{\mathbf{w} \in \Delta^{K-1}} R(\mathbf{w})$ for any $\mathbf{w}^* \in \mathcal{W}^*$. 

It follows that the event $\{d(\widehat{\mathbf{w}}_n, \mathcal{W}^*) \geq \varepsilon\}$ is a subset of the event $\{R(\widehat{\mathbf{w}}_n) - R(\mathbf{w}^*) \geq \eta\}$. Let $\rho_n:=\sup_{\mathbf w\in\Delta^{K-1}}|R_n(\mathbf w)-R(\mathbf w)|$, which converges to zero in probability by \textbf{uniform convergence} (Condition 1). By the definition of $\widehat{\mathbf w}_n$ as an approximate minimizer of $R_n$ (Ch. 5, \citealp{vaart:2000}), $R_n(\widehat{\mathbf w}_n)\le\inf_{\mathbf w\in\Delta^{K-1}}R_n(\mathbf w)+r_n$, with $r_n=o_p(1)$. Hence,
\[
R(\widehat{\mathbf w}_n)\le R_n(\widehat{\mathbf w}_n)+\rho_n\le R_n(\mathbf w^*)+r_n+\rho_n\le R(\mathbf w^*)+2\rho_n+r_n ,
\]
and, since $R(\widehat{\mathbf w}_n)\ge R(\mathbf w^*)$ by the definition of $\mathbf w^*$,
\[
0\le R(\widehat{\mathbf w}_n)-R(\mathbf w^*)\le2\rho_n+r_n\xrightarrow{\Pr}0 .
\]
Consequently,
\begin{equation*}
    \Pr\big(d(\widehat{\mathbf{w}}_n, \mathcal{W}^*) \geq \varepsilon\big) \leq \Pr\big(R(\widehat{\mathbf{w}}_n) - R(\mathbf{w}^*) \geq \eta\big) \leq \Pr\big(2\rho_n+r_n\geq\eta\big) \to 0, \quad \text{as } n \to \infty.
\end{equation*}
This concludes the proof.

\subsection*{Proof of Theorem~\ref{th:splitvalidity}} \label{proof:thm2}
Let $\mathcal T_n$ denote the $\sigma$-field generated by $\mathcal D_{\mathrm{tr}}$ and by the split and cross-validation randomization. 

Since the split is independent of the data, A0 implies that, conditionally on $\mathcal T_n$, the $n_{\mathrm{cal}}+1$ pairs in $\mathcal D_{\mathrm{cal}}\cup\{(\bX,Y)\}$ are exchangeable, and so are their non-conformity scores $s_{i,k}=\mathrm{score}\big(Y_i,\widehat f_k(\bX_i)\big)$, because $\widehat f_k$ is fixed given $\mathcal T_n$. Let $r=\lceil(1-\alpha)(n_{\mathrm{cal}}+1)\rceil$, where $n_{\mathrm{cal}}$ denotes the size of the calibration set ($\#\mathcal{I}_{\mathrm{cal}}$ in the Algorithm \ref{alg:splitCSL}). If $r>n_{\mathrm{cal}}$, then $\mathcal C_k(\bX;\mathcal D_n)=\mathbb R$ and $A^C_{n,k}=\emptyset$. Otherwise, $Y\notin\mathcal C_k(\bX;\mathcal D_n)$ if and only if the test score exceeds the $r$-th smallest calibration score, an event whose conditional probability is at most $(n_{\mathrm{cal}}+1-r)/(n_{\mathrm{cal}}+1)\le\alpha$ by exchangeability \citep{papadopoulos:2008,vovk:2022}. Hence
\begin{equation}\label{eq:trainvalid}
\Pr\big(A^C_{n,k}\mid\mathcal T_n\big)\le\alpha\quad\text{almost surely},\qquad k=1,\dots,K.
\end{equation}

\begin{itemize}
\item[(a)] The combined set fails to cover $Y$ if and only if $V_n^{\mathrm{mis}}\ge1/2$. Since the weights are fixed given $\mathcal T_n$, Markov's inequality and \eqref{eq:trainvalid} give
\[
\Pr\big(V_n^{\mathrm{mis}}\ge\tfrac12\mid\mathcal T_n\big)\le2\,\mathbb E\big[V_n^{\mathrm{mis}}\mid\mathcal T_n\big]=2\sum_{k=1}^K\widehat w_{n,k}\Pr\big(A^C_{n,k}\mid\mathcal T_n\big)\le2\alpha .
\]
Taking expectations and using the tower rule of expectations (law of total expectation) gives the unconditional bound.

\item[(b)] On $\Omega_n$, the maximizer $\widehat k:=\arg\max_k\widehat w_{n,k}$ is unique and
$\mathcal C_{\mathrm{comb}}=\mathcal C_{\widehat k}(\bX;\mathcal D_n)$ (Section~\ref{sec:theory}).
Since $\Omega_n$ and $\widehat k$ are $\mathcal T_n$-measurable,
$\mathbf 1_{\Omega_n}\Pr\big(Y\notin\mathcal C_{\mathrm{comb}}\mid\mathcal T_n\big)
=\sum_{k=1}^K\mathbf 1_{\Omega_n}\mathbf 1_{\{\widehat k=k\}}\Pr\big(A^C_{n,k}\mid\mathcal T_n\big)
\le\alpha\mathbf 1_{\Omega_n}$, while (a) gives
$\mathbf 1_{\Omega_n^C}\Pr\big(Y\notin\mathcal C_{\mathrm{comb}}\mid\mathcal T_n\big)\le2\alpha\mathbf 1_{\Omega_n^C}$.
Adding the two bounds,
$\Pr\big(Y\notin\mathcal C_{\mathrm{comb}}\mid\mathcal T_n\big)\le\alpha\mathbf 1_{\Omega_n}+2\alpha\mathbf 1_{\Omega_n^C}$
almost surely.
Taking expectations leads to $\Pr\big(Y\notin\mathcal C_{\mathrm{comb}}(\bX;\mathcal D_n)\big)\le\alpha\Pr(\Omega_n)+2\alpha\Pr(\Omega_n^C)=\alpha+\alpha\Pr(\Omega_n^C)$.

Finally, suppose that A2 holds and let $\gamma:=\inf_{\mathbf w\in\mathcal W^*}w_{k^\star}-\tfrac12>0$. If $d(\widehat{\mathbf w}_n,\mathcal W^*)<\gamma$, there exists $\mathbf w\in\mathcal W^*$ with $\|\widehat{\mathbf w}_n-\mathbf w\|<\gamma$, so that $\widehat w_{n,k^\star}>w_{k^\star}-\gamma\ge\tfrac12$ and $\Omega_n$ occurs. Hence $\Pr(\Omega_n^C)\le\Pr\big(d(\widehat{\mathbf w}_n,\mathcal W^*)\ge\gamma\big)\to0$ by A2 (applied to the weights computed from $\mathcal D_{\mathrm{tr}}$, with $n_{\mathrm{tr}}\to\infty$), and the last statement follows from the bound just obtained.

\item[(c)] Recall that conditioning on $\mathcal D_{\mathrm{tr}}$ includes conditioning on the
split and cross-validation randomization, that is, on $\mathcal T_n$. Let $\widehat j$ be the
smallest index in $\widehat{\mathcal K}$ attaining the minimum in the definition of
$\widehat{\zeta}_n$. Since $\widehat{\mathcal K}$ is $\mathcal T_n$-measurable and the
conditional probabilities defining $\widehat{\zeta}_n$ are $\mathcal T_n$-measurable, so is
$\widehat j$.

On $\Omega_n^{\widehat{\mathcal K}}$, if every learner in $\widehat{\mathcal K}$ covers $Y$, the
vote total for $Y$ is at least $\sum_{k\in\widehat{\mathcal K}}\widehat w_{n,k}>1/2$, so that
$Y\in\mathcal C_{\mathrm{comb}}(\bX;\mathcal D_n)$. Hence, on $\Omega_n^{\widehat{\mathcal K}}$,
\[
\{Y\notin\mathcal C_{\mathrm{comb}}\}\subseteq\bigcup_{k\in\widehat{\mathcal K}}A^C_{n,k}
\subseteq A^C_{n,\widehat j}\cup\bigcup_{k\in\widehat{\mathcal K}}
\big(A^C_{n,\widehat j}\triangle A^C_{n,k}\big).
\]
By the union bound, and since $\widehat j$ is $\mathcal T_n$-measurable, so that
$\Pr(A^C_{n,\widehat j}\mid\mathcal T_n)=\sum_{j}\mathbf 1_{\{\widehat j=j\}}
\Pr(A^C_{n,j}\mid\mathcal T_n)\le\alpha$ by \eqref{eq:trainvalid},
\[
\Pr\big(Y\notin\mathcal C_{\mathrm{comb}}\mid\mathcal T_n\big)\le\alpha
+\sum_{k\in\widehat{\mathcal K}}\Pr\big(A^C_{n,\widehat j}\triangle A^C_{n,k}\mid\mathcal T_n\big)
=\alpha+\widehat{\zeta}_n\quad\text{on }\Omega_n^{\widehat{\mathcal K}}.
\]
Combining this with (a),
$\Pr(Y\notin\mathcal C_{\mathrm{comb}}\mid\mathcal T_n)\le\{\alpha+\min(\widehat{\zeta}_n,\alpha)\}
\mathbf 1_{\Omega_n^{\widehat{\mathcal K}}}+2\alpha\mathbf 1_{(\Omega_n^{\widehat{\mathcal K}})^C}$
almost surely, and taking expectations gives
\[
\Pr\big(Y\notin\mathcal C_{\mathrm{comb}}\big)\le\alpha+\alpha\Pr\big((\Omega_n^{\widehat{\mathcal K}})^C\big)
+\mathbb E\big[\min(\widehat{\zeta}_n,\alpha)\mathbf 1_{\Omega_n^{\widehat{\mathcal K}}}\big],
\]
which is the stated bound. If $\widehat{\mathcal K}=\{\widehat k\}$, the sum defining
$\widehat{\zeta}_n$ contains only the term $k=\widehat k$, which is zero.

Finally, let $G_n:=\{\widehat{\mathcal K}=\mathcal K^*\}$, so that $\Pr(G_n^C)\to0$, and fix
$i\in\mathcal K^*$. On $G_n$,
$\Omega_n^{\widehat{\mathcal K}}=\{\sum_{k\in\mathcal K^*}\widehat w_{n,k}>\tfrac12\}$, so that
\[
\Pr\big((\Omega_n^{\widehat{\mathcal K}})^C\big)\le\Pr(G_n^C)
+\Pr\Big(\sum_{k\in\mathcal K^*}\widehat w_{n,k}\le\tfrac12\Big)\to0,
\]
where the second term vanishes by the argument in the proof of
Theorem~\ref{th:marginalvalidity}(c), applied to the weights computed from
$\mathcal D_{\mathrm{tr}}$ with $n_{\mathrm{tr}}\to\infty$. Moreover, on $G_n$ the minimum
defining $\widehat{\zeta}_n$ is over $\mathcal K^*$, so
$\widehat{\zeta}_n\le\sum_{k\in\mathcal K^*}\Pr\big(A^C_{n,i}\triangle A^C_{n,k}\mid\mathcal T_n\big)$,
and therefore
$\min(\widehat{\zeta}_n,\alpha)\mathbf 1_{\Omega_n^{\widehat{\mathcal K}}}\le\alpha\mathbf 1_{G_n^C}
+\sum_{k\in\mathcal K^*}\Pr\big(A^C_{n,i}\triangle A^C_{n,k}\mid\mathcal T_n\big)$.
Taking expectations and using the tower property,
\[
\mathbb E\big[\min(\widehat{\zeta}_n,\alpha)\mathbf 1_{\Omega_n^{\widehat{\mathcal K}}}\big]
\le\alpha\Pr(G_n^C)+\sum_{k\in\mathcal K^*}\Pr\big(A^C_{n,i}\triangle A^C_{n,k}\big)\to0 ,
\]
and the last statement follows from the bound just obtained.
\end{itemize}

\subsection*{Proof of Theorem~\ref{th:marginalvalidity}}
\begin{itemize}
\item[(a)] Since $\widehat{\mathbf w}_n$ is $\mathcal D_n$-measurable, $\mathbb E\big[V_n^{\mathrm{mis}}\mid\mathcal D_n\big]=\sum_k\widehat w_{n,k}P_{n,k}\le\max_kP_{n,k}\le\alpha+\max_k(P_{n,k}-\alpha)_+$. By Markov's inequality,
\[
\Pr\big(Y\notin\mathcal C_{\mathrm{comb}}(\bX;\mathcal D_n)\big)=\Pr\big(V_n^{\mathrm{mis}}\ge\tfrac12\big)\le2\,\mathbb E\big[V_n^{\mathrm{mis}}\big]\le2\alpha+2\,\mathbb E\Big[\max_k(P_{n,k}-\alpha)_+\Big].
\]
Under A1$^\star$, for every $\varepsilon>0$, $\Pr\big(\max_k(P_{n,k}-\alpha)_+>\varepsilon\big)\le\sum_k\Pr(P_{n,k}>\alpha+\varepsilon)\to0$. Since $\max_k(P_{n,k}-\alpha)_+\in[0,1]$, bounded convergence gives $\mathbb E\big[\max_k(P_{n,k}-\alpha)_+\big]\to0$, and hence $\limsup_n\Pr\big(Y\notin\mathcal C_{\mathrm{comb}}(\bX;\mathcal D_n)\big)\le2\alpha$.

\item[(b)] Let $\gamma:=\inf_{\mathbf w\in\mathcal W^*}w_{k^\star}-\tfrac12>0$. If $d(\widehat{\mathbf w}_n,\mathcal W^*)<\gamma$, there exists $\mathbf w\in\mathcal W^*$ with $\|\widehat{\mathbf w}_n-\mathbf w\|<\gamma$, so that $\widehat w_{n,k^\star}>w_{k^\star}-\gamma\ge\tfrac12$. Hence, by A2,
\[
\Pr\big(\widehat w_{n,k^\star}\le\tfrac12\big)\le\Pr\big(d(\widehat{\mathbf w}_n,\mathcal W^*)\ge\gamma\big)\to0 .
\]
On the event $\{\widehat w_{n,k^\star}>1/2\}$, $\mathcal C_{\mathrm{comb}}(\bX;\mathcal D_n)=\mathcal C_{k^\star}(\bX;\mathcal D_n)$. Therefore,
\[
\Pr\big(Y\notin\mathcal C_{\mathrm{comb}}(\bX;\mathcal D_n)\big)\le\Pr\big(A^C_{n,k^\star}\big)+\Pr\big(\widehat w_{n,k^\star}\le\tfrac12\big),
\]
and the limit superior of the right-hand side is at most $\alpha$ by A1.

\item[(c)] Let $W_n:=\sum_{k\in\mathcal K^*}\widehat w_{n,k}$ and $E_n:=\bigcup_{k\in\mathcal K^*}A^C_{n,k}$. On $E_n^C$, every learner in $\mathcal K^*$ covers $Y$, so the vote total for $Y$ is at least $W_n$. Hence $E_n^C\cap\{W_n>\tfrac12\}\subseteq\{Y\in\mathcal C_{\mathrm{comb}}(\bX;\mathcal D_n)\}$, and
\[
\Pr\big(Y\notin\mathcal C_{\mathrm{comb}}(\bX;\mathcal D_n)\big)\le\Pr(E_n)+\Pr\big(W_n\le\tfrac12\big).
\]
If $d(\widehat{\mathbf w}_n,\mathcal W^*)<\delta/\sqrt K$, there exists $\mathbf w\in\mathcal W^*$ with $\|\widehat{\mathbf w}_n-\mathbf w\|<\delta/\sqrt K$, and, by the Cauchy--Schwarz inequality,
\[
W_n\ge\sum_{k\in\mathcal K^*}w_k-\sqrt K\,\|\widehat{\mathbf w}_n-\mathbf w\|>\tfrac12 .
\]
Hence, by A2, $\Pr(W_n\le\tfrac12)\le\Pr\big(d(\widehat{\mathbf w}_n,\mathcal W^*)\ge\delta/\sqrt K\big)\to0$. Moreover, fixing $j\in\mathcal K^*$, we have $E_n\subseteq A^C_{n,j}\cup\bigcup_{k\in\mathcal K^*}\big(A^C_{n,j}\,\triangle\,A^C_{n,k}\big)$, so that
\[
\Pr(E_n)\le\Pr\big(A^C_{n,j}\big)+\sum_{k\in\mathcal K^*}\Pr\big(A^C_{n,j}\,\triangle\,A^C_{n,k}\big),
\]
whose limit superior is at most $\alpha$ by A1 and the agreement condition. The case $\mathcal K^*=\{k^\star\}$ follows with $\delta=\inf_{\mathbf w\in\mathcal W^*}w_{k^\star}-\tfrac12$.
\end{itemize}

\subsection*{Proof of Theorem~\ref{th:conditionalvalidity}}
Fix $\bx\in\mathcal X$. Since $(\bX,Y)$ is independent of $\mathcal D_n$, the conditional distribution of $\mathcal D_n$ given $\bX=\bx$ coincides with its unconditional distribution. Consequently, $\Pr(E\mid\bX=\bx)=\Pr(E)$ for any event $E$ that depends on $\mathcal D_n$ only. Moreover, since $\widehat{\mathbf w}_n$ is $\mathcal D_n$-measurable and $\Pr\big(A^C_{n,k}\mid\mathcal D_n,\bX=\bx\big)=P_{n,k}(\bx)$,
\[
\mathbb E\big[V_n^{\mathrm{mis}}\mid\bX=\bx\big]=\mathbb E\Big[\sum_{k=1}^K\widehat w_{n,k}P_{n,k}(\bx)\Big].
\]
\begin{itemize}
\item[(a)] By Markov's inequality conditionally on $\bX=\bx$,
\[
\Pr\big(Y\notin\mathcal C_{\mathrm{comb}}(\bX;\mathcal D_n)\mid\bX=\bx\big)\le2\,\mathbb E\big[V_n^{\mathrm{mis}}\mid\bX=\bx\big]\le2\alpha+2\,\mathbb E\Big[\max_k\big(P_{n,k}(\bx)-\alpha\big)_+\Big],
\]
and the last term vanishes under A1$^\star(\bx)$ by bounded convergence, as in the proof of Theorem~\ref{th:marginalvalidity}(a).

\item[(b)] As in the proof of Theorem~\ref{th:marginalvalidity}(b),
\[
\Pr\big(Y\notin\mathcal C_{\mathrm{comb}}(\bX;\mathcal D_n)\mid\bX=\bx\big)\le\Pr\big(A^C_{n,k^\star}\mid\bX=\bx\big)+\Pr\big(\widehat w_{n,k^\star}\le\tfrac12\big),
\]
where we have used $\Pr\big(\widehat w_{n,k^\star}\le\tfrac12\mid\bX=\bx\big)=\Pr\big(\widehat w_{n,k^\star}\le\tfrac12\big)$. The first term has limit superior at most $\alpha$ by A1$(\bx)$, and the second tends to zero by A2.

\item[(c)] With $W_n$ and $E_n$ as in the proof of Theorem~\ref{th:marginalvalidity}(c), the same argument gives
\[
\Pr\big(Y\notin\mathcal C_{\mathrm{comb}}(\bX;\mathcal D_n)\mid\bX=\bx\big)\le\Pr\big(E_n\mid\bX=\bx\big)+\Pr\big(W_n\le\tfrac12\big),
\]
where we have used $\Pr\big(W_n\le\tfrac12\mid\bX=\bx\big)=\Pr\big(W_n\le\tfrac12\big)$, since $W_n$ depends on $\mathcal D_n$ only. The second term tends to zero by A2, as shown in that proof. Fixing $j\in\mathcal K^*$,
\[
\Pr\big(E_n\mid\bX=\bx\big)\le\Pr\big(A^C_{n,j}\mid\bX=\bx\big)+\sum_{k\in\mathcal K^*}\Pr\big(A^C_{n,j}\,\triangle\,A^C_{n,k}\mid\bX=\bx\big),
\]
whose limit superior is at most $\alpha$ by A1$(\bx)$ and the conditional agreement condition.
\end{itemize}

\subsection*{Computing times}

All CSL prediction intervals reported in this work are constructed using a grid resolution of 0.0001 for evaluating candidate response values. All simulations were conducted in R version 4.5.2 (2025-10-31) on a machine running macOS (Darwin), equipped with an Apple M2 Pro processor, 16 GB of RAM, and using a single CPU core.

\begin{table}[!htbp]
\centering
\begin{tabular}{llcc}
\toprule
\textbf{Scenario} & \textbf{Sample Size} & \textbf{Split Conformal} & \textbf{Full Conformal} \\
\midrule
\multirow{3}{*}{\textbf{S1}} 
 & n = 100  & 0.137/0.035 & 0.671/0.263 \\
 & n = 500  & 0.409/0.036 & 1.922/1.668 \\
 & n = 1000 & 0.874/0.056 & 4.148/3.735 \\
\midrule
\multirow{3}{*}{\textbf{S2}} 
 & n = 100  & 0.143/0.043 & 0.636/0.083 \\
 & n = 500  & 0.435/0.017 & 1.861/0.942 \\
 & n = 1000 & 0.943/0.032 & 4.136/0.824 \\
\midrule
\multirow{3}{*}{\textbf{S3}} 
 & n = 100  & 0.161/0.075 & 0.712/0.177 \\
 & n = 500  & 0.322/0.039 & 1.551/0.858 \\
 & n = 1000 & 0.630/0.086 & 3.451/2.438 \\
\midrule
\multirow{3}{*}{\textbf{S4}} 
 & n = 100  & 0.845/0.513 & 3.628/1.286 \\
 & n = 500  & 1.118/0.120 & 5.712/2.613 \\
 & n = 1000 & 2.017/0.151 & 10.918/7.626 \\
\bottomrule
\end{tabular}
\caption{\textbf{CPU time (in seconds) and standard deviation (Average Time / Standard Deviation) for each scenario under different sample sizes.}}
\label{tab:simulation_time}
\end{table}

Table \ref{tab:simulation_time} presents the average CPU time (in seconds) and its standard deviation across 100 iterations for each simulation scenario and sample size, corresponding to the settings in Table \ref{tab:coverage}. As expected, the full conformal procedure is substantially more expensive than the split conformal procedure, and computing times increase with both sample size and covariate dimension, with scenario S4 being the most demanding due to its higher-dimensional covariate space.

For the case study, computations were carried out on the CREATE high-performance computing cluster at King's College London using Slurm batch jobs in R (version 4.5.1). We requested 1 node, 1 task, 8 CPU cores, and 32 GB of memory from the CPU partition. The application involved two main computational tasks. First, constructing Full-CSL prediction intervals for the testing set: with 502 testing observations in the main analysis (492 in the sensitivity analysis), approximately 25 hours (23 hours, respectively) were required per 100 testing points. Second, constructing covariate-specific prediction interval curves by varying each of the 12 continuous covariates over a grid of 15 values across three representative profiles, yielding 45 Full-CSL intervals per variable; each variable required approximately 11.5 hours in the main analysis and 10.5 hours in the sensitivity analysis. By contrast, the Split-CSL procedure is computationally inexpensive: for the first task, the total time was less than 300 seconds; for the second task, a finer grid of 30 values per covariate was used across the same three representative profiles, and the total time remained similarly negligible.

\subsection*{Appendix for the simulation study}

For Scenario S4 with sample size \(n=100\), 20 out of the 1000 Monte Carlo replications were excluded due to non-convergence of the GAM fitting routine in \texttt{mgcv}. In all skipped cases, the same error message was returned: ``magic, the gcv/ubre optimizer, failed to converge after 400 iterations.'' The skipped seeds were
\[
17,\,147,\,226,\,243,\,291,\,353,\,365,\,408,\,413,\,618,\,655,\,670,\,731,\,745,\,774,\,813,\,866,\,875,\,924,\,937.
\]
Therefore, the reported empirical coverage and average interval width for this setting were computed from the remaining 980 successful replications.

\subsection*{Appendix for the case study}

The data set is constructed by merging 10 NHANES modules using the participant identifier \texttt{SEQN}: the demographic file \texttt{DEMO\_L}, the body measures file \texttt{BMX\_L}, the biochemistry profile file \texttt{BIOPRO\_L}, the triglycerides file \texttt{TRIGLY\_L}, the medical conditions file \texttt{MCQ\_L}, the diabetes questionnaire file \texttt{DIQ\_L}, the glycohemoglobin file \texttt{GHB\_L}, the blood pressure questionnaire file \texttt{BPQ\_L}, the prescription medication file \texttt{RXQ\_RX\_L}, and the blood pressure file \texttt{BPXO\_L}. Given its close relationship with creatinine levels, the kidney conditions--urology module \texttt{KIQ\_U\_L} is used to exclude participants based on the dialysis-related variable \texttt{KIQ025}, although this module is not merged into the final data set. Pregnant participants are also excluded based on the pregnancy status variable \texttt{RIDEXPRG} from \texttt{DEMO\_L}. 

The continuous covariates, summarized in the form $(\min,\ \mathrm{mean},\ \max)$, are as follows: age, \texttt{RIDAGEYR} (years), $(20.0, 54.03, 80.0)$; body mass index, \texttt{BMXBMI} (kg/m$^2$), $(14.90, 29.63, 68.90)$; waist circumference, \texttt{BMXWAIST} (cm), $(62.40, 101.00, 187.00)$; systolic blood pressure (first oscillometric reading), \texttt{BPXOSY1}, $(75.00, 122.80, 225.00)$; albumin, \texttt{LBXSAL} (g/dL), $(2.20, 4.08, 5.50)$; total protein, \texttt{LBXSTP} (g/dL), $(5.00, 7.14, 9.00)$; uric acid, \texttt{LBXSUA} (mg/dL), $(1.80, 5.17, 13.20)$; triglycerides, \texttt{LBXSTR} (mg/dL), $(24.0, 134.7, 1724.0)$; total cholesterol, \texttt{LBXSCH} (mg/dL), $(63.0, 190.3, 454.0)$; serum calcium, \texttt{LBXSCA} (mg/dL), $(7.50, 9.41, 11.70)$; serum phosphorus, \texttt{LBXSPH} (mg/dL), $(1.60, 3.51, 5.70)$; and glycohemoglobin, \texttt{LBXGH} (\%), $(3.20, 5.78, 17.10)$. The categorical covariates are summarized by category frequencies as follows: sex, \texttt{RIAGENDR}, with 2304 males and 2723 females; race/ethnicity, \texttt{RIDRETH3}, with 336 Mexican American, 509 Other Hispanic, 3058 Non-Hispanic White, 549 Non-Hispanic Black, 267 Non-Hispanic Asian, and 308 Other Race, including Multi-Racial participants; liver disease status, \texttt{MCQ160L}, with 285 ``Yes'' and 4742 ``No'' responses; diabetes status, \texttt{DIQ010}, with 680 ``Yes'', 4170 ``No'', and 177 ``Borderline'' responses; high blood pressure status, \texttt{BPQ020}, with 1873 ``Yes'' and 3154 ``No'' responses; and prescription medication use, \texttt{RXQ033}, with 3530 ``Yes'' and 1497 ``No'' responses. Pairwise collinearity among the continuous covariates are assessed using the Pearson correlation matrix and scatter plots. Apart from the strong association between \texttt{BMXBMI} and \texttt{BMXWAIST} (0.895), the remaining correlations are less than 0.5, so both variables were retained given their clinical relevance and the predictive focus of the analysis.

\begin{figure}[!htbp]
  \centering
  \includegraphics[width=0.9\textwidth]{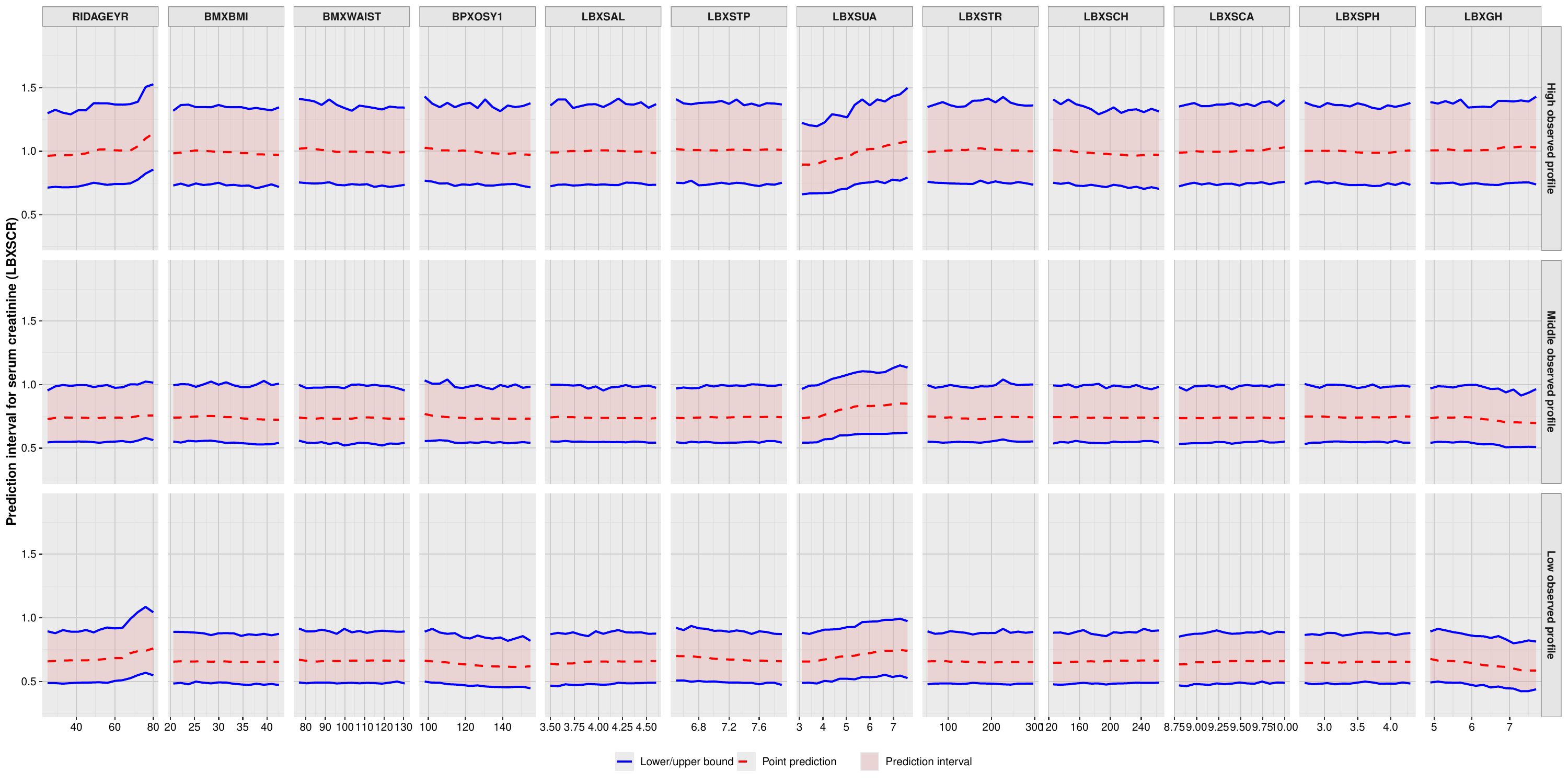}
  \caption{\textbf{Covariate-specific full-CSL prediction intervals for three representative profiles.}}
  \label{fig:prediction_intervals_covariate}
\end{figure}

Figure \ref{fig:prediction_intervals_covariate} shows covariate-specific Full-CSL prediction interval curves for serum creatinine based on three representative testing profiles. These profiles are chosen from observations whose observed responses lie close to the 90th, 50th, and 10th percentiles of the ordered test responses, corresponding to the high, middle, and low profiles shown in the first, second, and third rows, respectively. In each panel, one covariate is varied over its empirical 5th to 95th percentile range using 15 equally spaced grid points, while all other covariates are kept fixed at the selected profile values (for age, \texttt{RIDAGEYR}, values are rounded to integers). The shaded region shows the Full-CSL prediction interval, the solid lines show its lower and upper bounds, and the dashed line shows the SL point prediction. The figure is used to examine how the predicted level of serum creatinine and the associated uncertainty change across the covariate space for subjects with different baseline profiles.
Several patterns emerge from Figure \ref{fig:prediction_intervals_covariate}. Uric acid (\texttt{LBXSUA}) exhibits the clearest positive association with predicted serum creatinine: across all three profiles, both the point prediction and interval bounds increase steadily as uric acid rises. Age (\texttt{RIDAGEYR}) also shows a positive trend, most clearly under the high and low profiles, while the effect is weaker for the middle profile. Blood pressure (\texttt{BPXOSY1}) displays a negative association, with predicted creatinine declining as blood pressure increases. The remaining covariates produce mostly flat curves, suggesting that their individual contributions to predicted creatinine are limited when other covariates are held fixed.
The figure is also informative regarding predictive uncertainty. For many covariates, interval width remains fairly stable, suggesting that uncertainty changes little across much of the covariate space. However, the intervals widen noticeably in panels where the point prediction varies more sharply, most clearly for age and uric acid. This indicates greater uncertainty in regions associated with larger predicted creatinine values. More broadly, the flatness of many panels suggests that the fitted Full-CSL structure is driven primarily by a small subset of influential variables rather than by all covariates equally.

\begin{figure}[!htbp]
  \centering
  \includegraphics[width=0.42\textwidth]{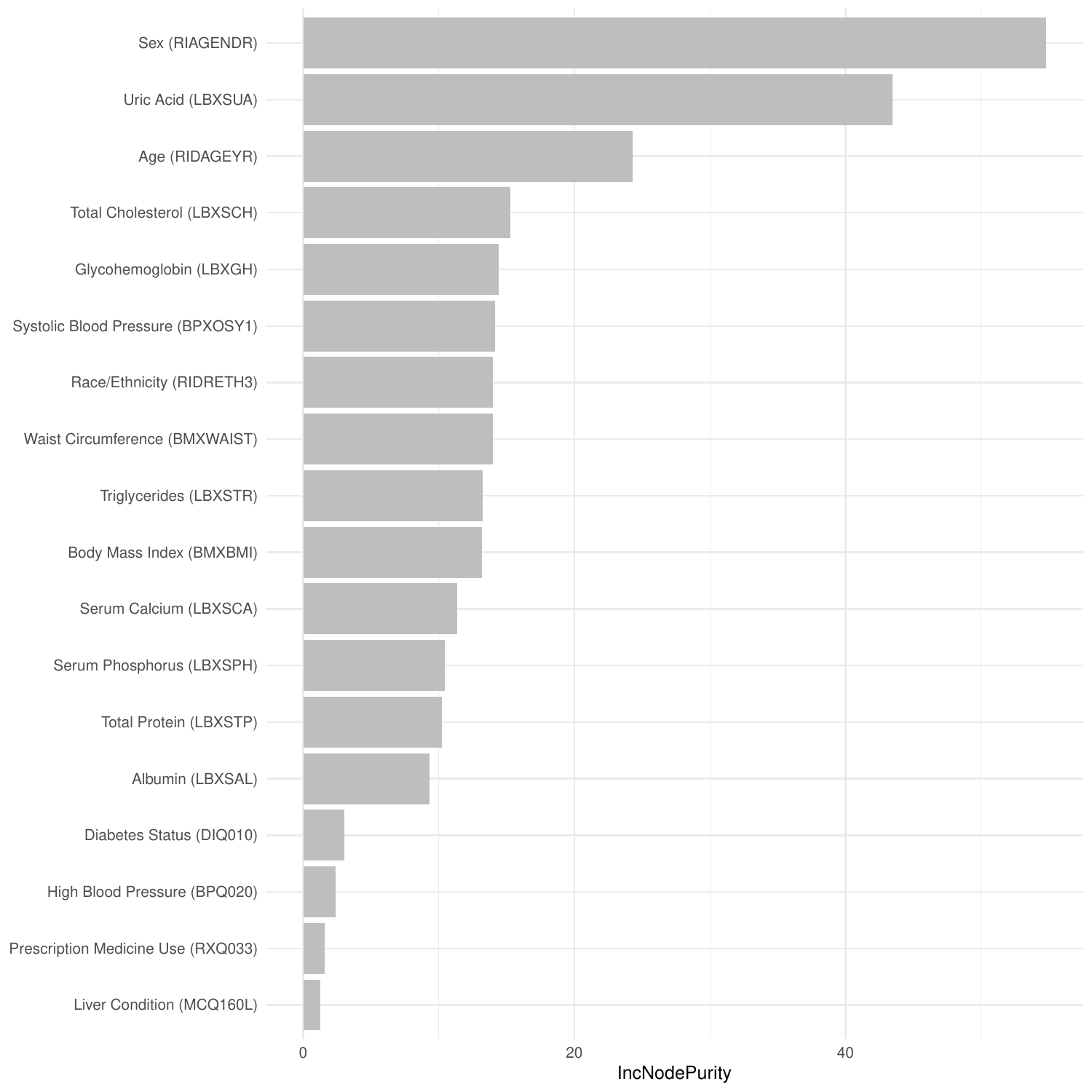}
  \hfill
  \includegraphics[width=0.42\textwidth]{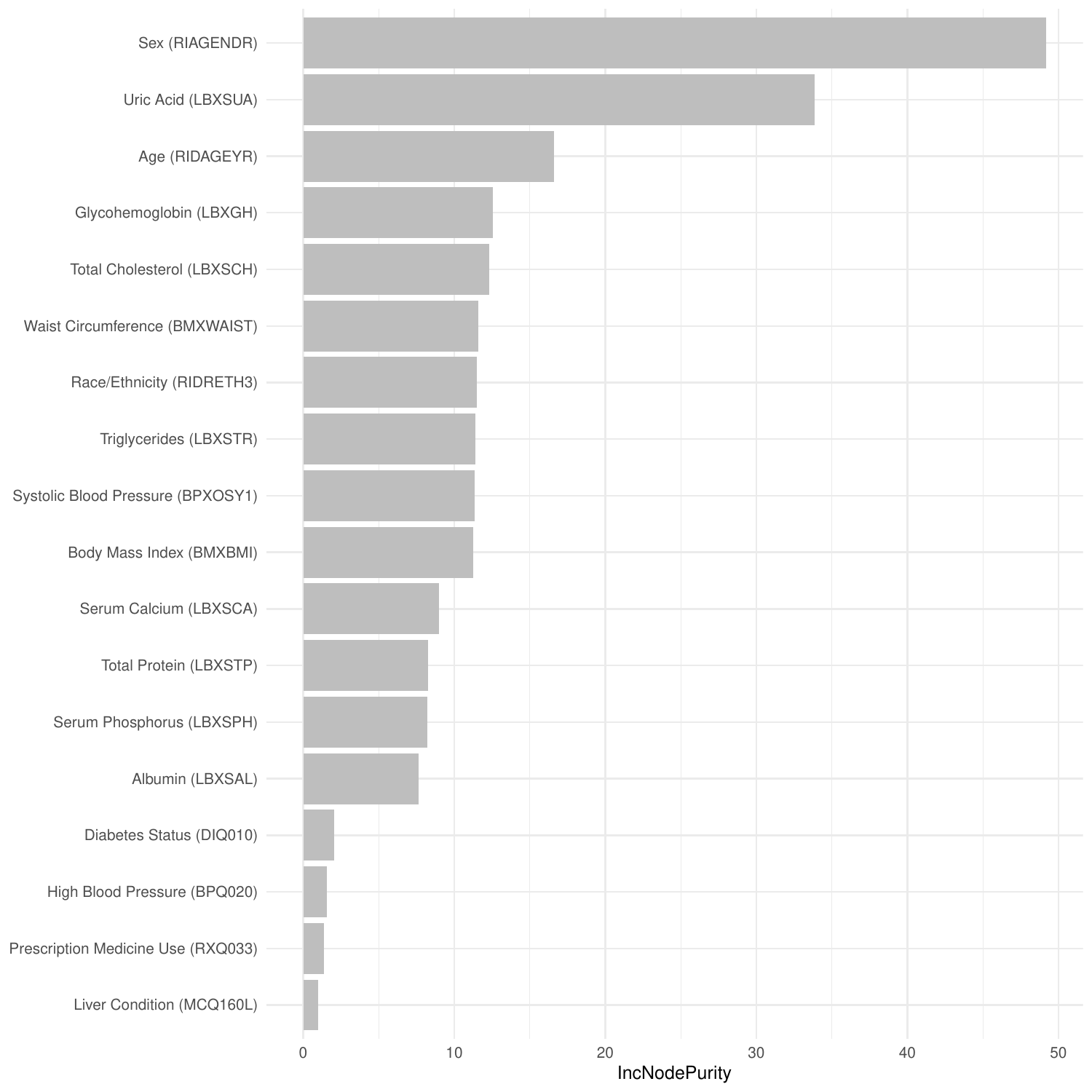}
  \caption{\textbf{Random Forest variable importance (IncNodePurity) for predicting serum creatinine. Left: full data set; Right: after removing observations with serum creatinine > 1.5 mg/dL.}}
  \label{fig:variable_importance}
\end{figure}

\begin{figure}[!htbp]
  \centering
  \includegraphics[width=0.7\textwidth]{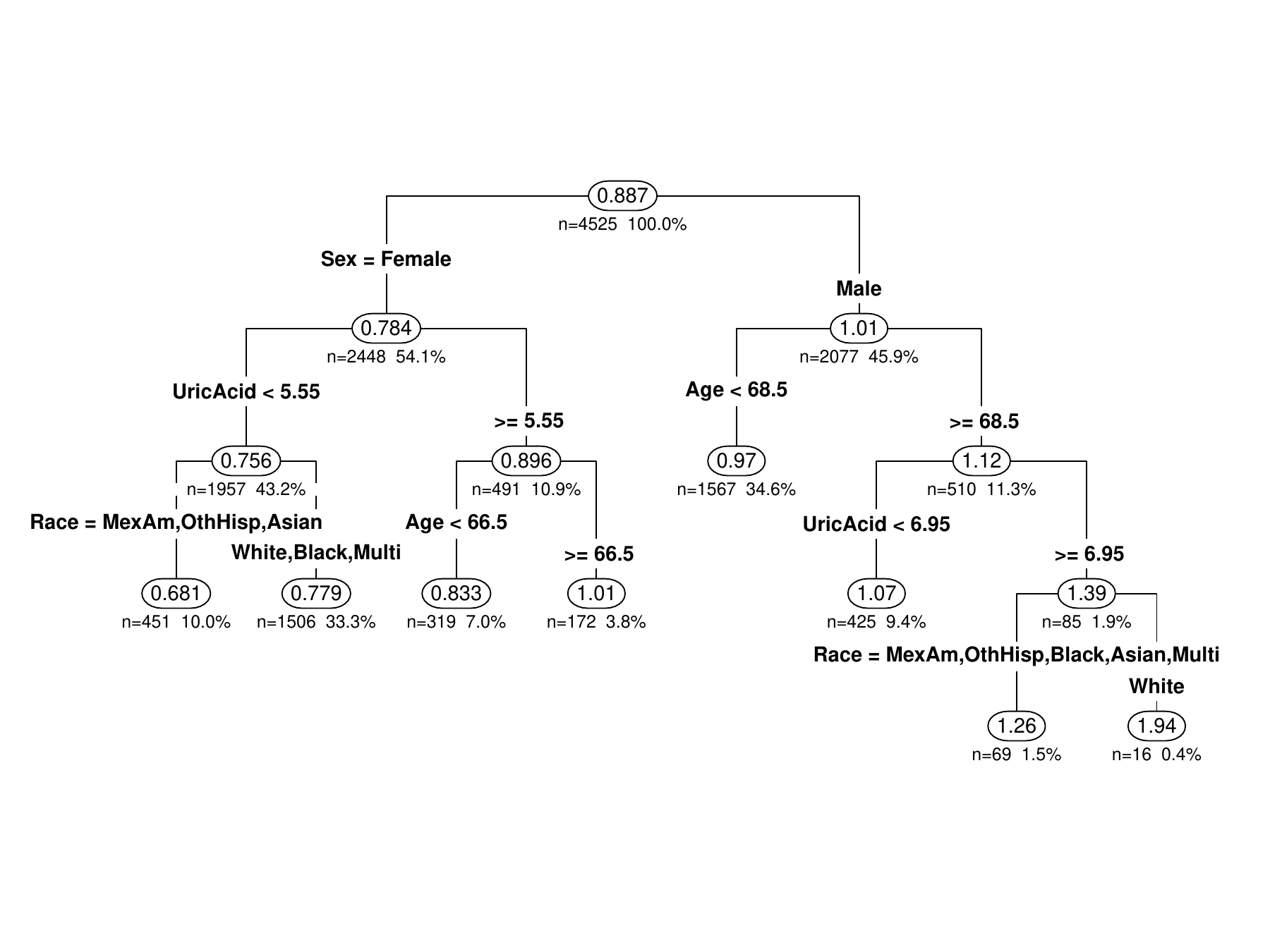}
  \caption{\textbf{Regression tree fitted as an interpretable surrogate for the dominant Random Forest learner.}}
  \label{fig:treeplot_not}
\end{figure}

\begin{figure}[!htbp]
  \centering
  \includegraphics[width=0.7\textwidth]{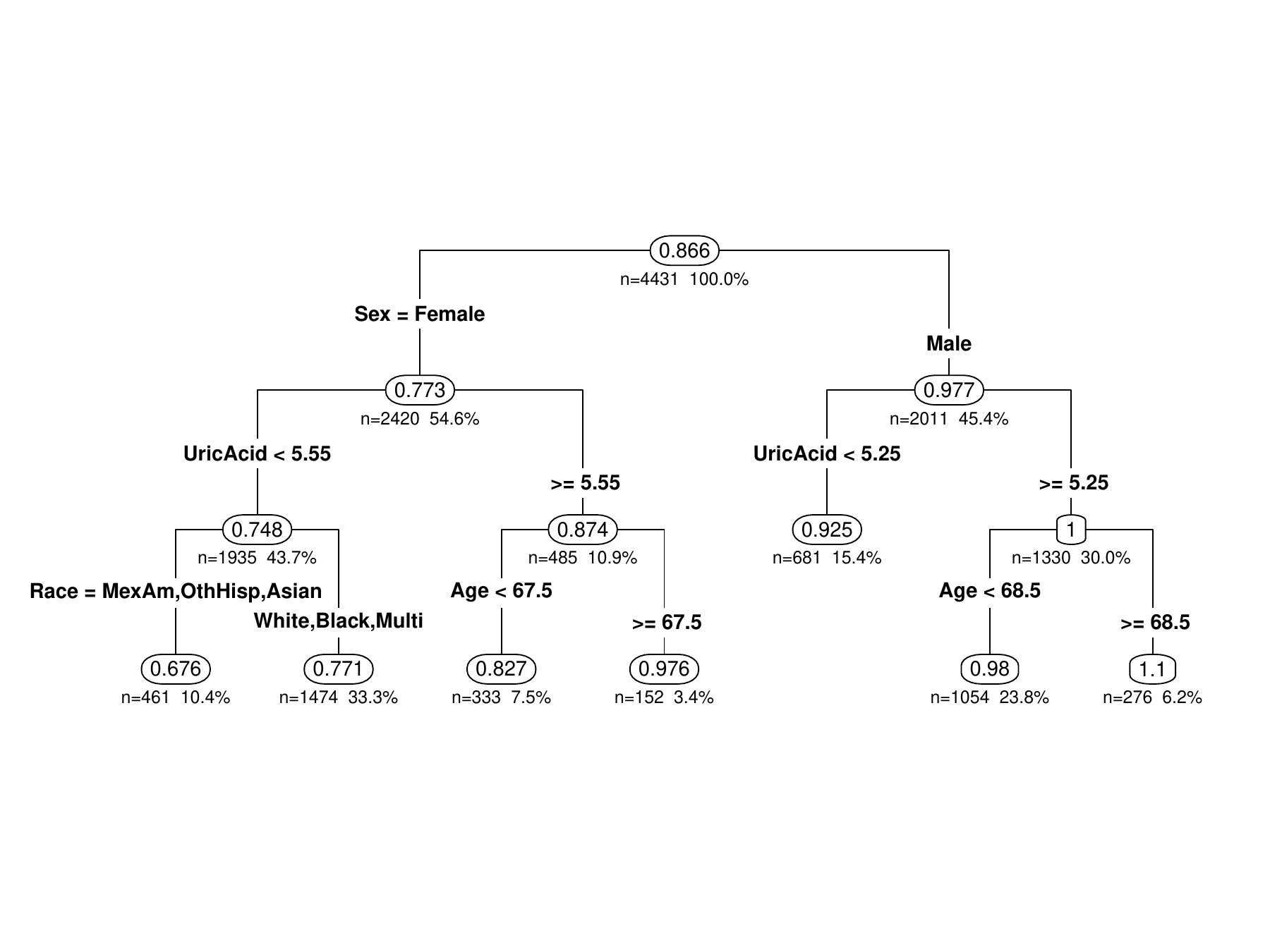}
  \caption{\textbf{Regression tree fitted as an interpretable surrogate for the dominant Random Forest learner after removing outliers.}}
  \label{fig:treeplot_rem}
\end{figure}

Given that Random Forest receives the dominant weight under both conformal settings, we examine its variable importance and approximate its splitting structure using an interpretable surrogate. Figure \ref{fig:variable_importance} displays variable importance, measured by Increase in Node Purity. The left and right panels correspond to the full data set and the outlier-removed data set, respectively. In both settings, sex, uric acid, and age rank as the three most influential predictors. To obtain an interpretable approximation of the RF, we fit regression trees using the \texttt{rpart} package. Figures \ref{fig:treeplot_not} and \ref{fig:treeplot_rem} display the resulting trees for the full data set and the outlier-removed data set, respectively. Node values represent predicted serum creatinine (mg/dL), and percentages indicate the proportion of observations. In both trees, sex is the primary split: females have lower predicted creatinine than males. Subsequent splits involve uric acid and age, consistent with the importance rankings in Figure \ref{fig:variable_importance}. In the full data set, predicted values range from 0.681 mg/dL to 1.94 mg/dL; after outlier removal, values range from 0.676 mg/dL to 1.10 mg/dL. Race/ethnicity contributes to further stratification within subgroups. The surrogate trees confirm that sex, uric acid, and age are the primary drivers of predicted serum creatinine. Additionally, the tree fitted after outlier removal is shallower, reflecting the reduced variability in the response and the simpler predictive structure required to capture the remaining observations.

\begin{figure}[!htbp]
  \centering
  \includegraphics[width=0.42\textwidth]{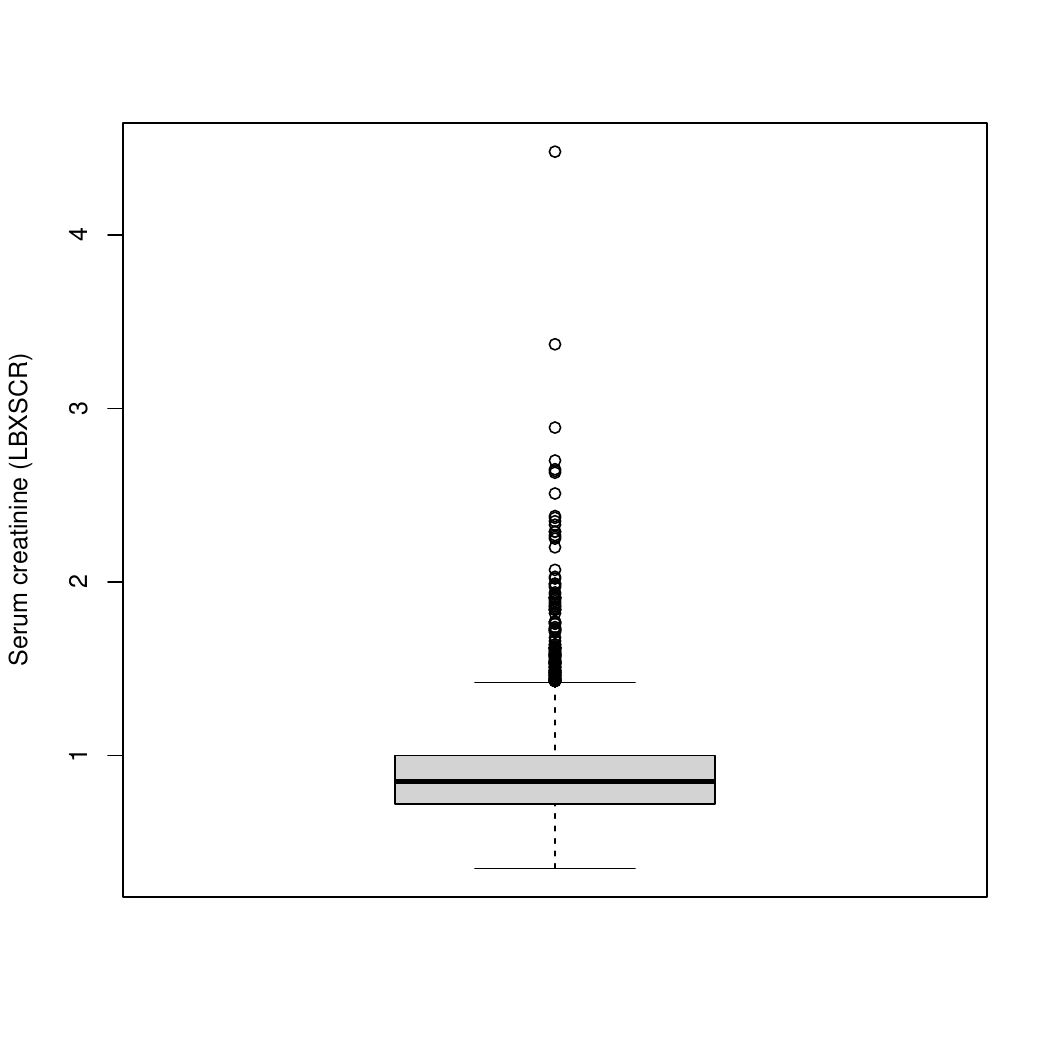}
  \hfill
  \includegraphics[width=0.42\textwidth]{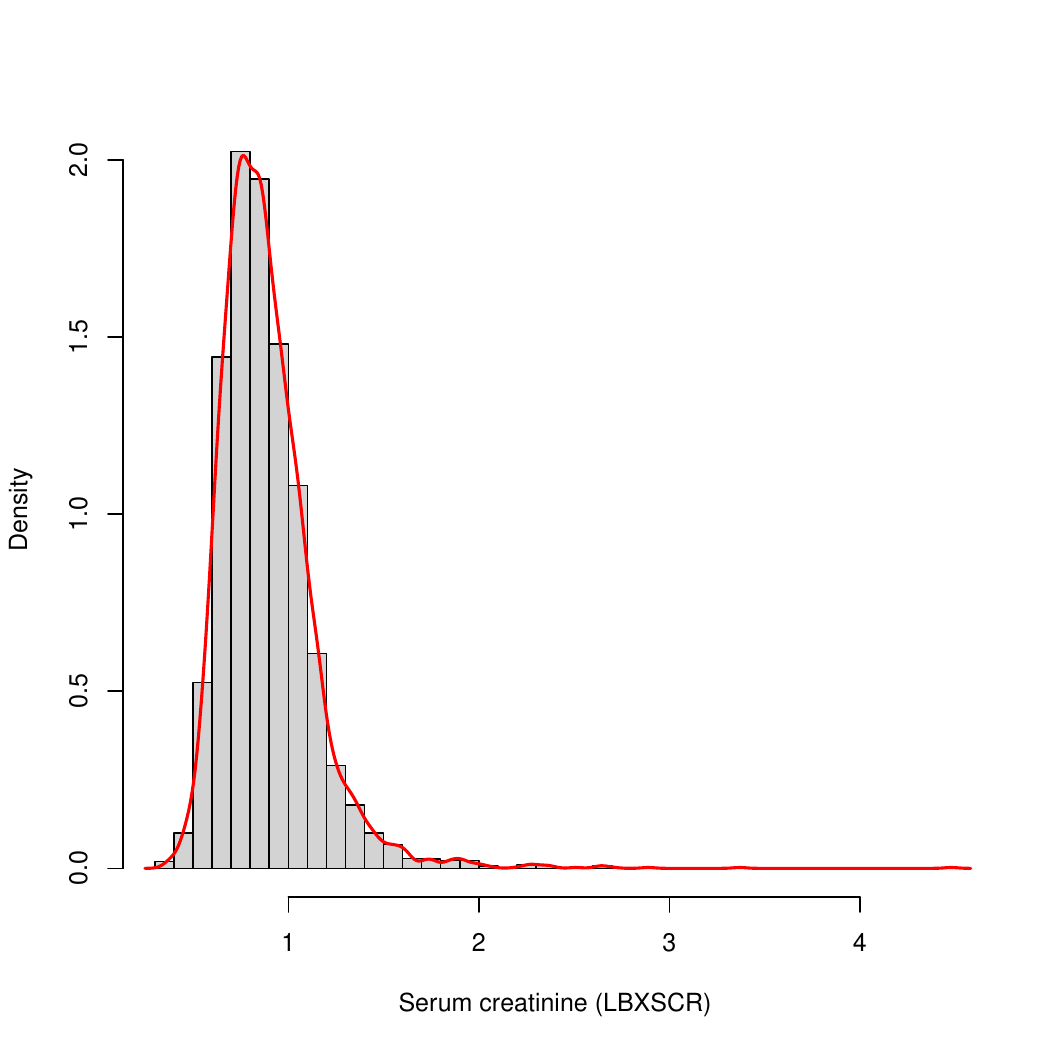}
  \caption{\textbf{Boxplot and histogram for the response (LBXSCR).}}
  \label{fig:response_plots}
\end{figure}


\begin{figure}[!htbp]
  \centering
  \includegraphics[width=0.9\textwidth]{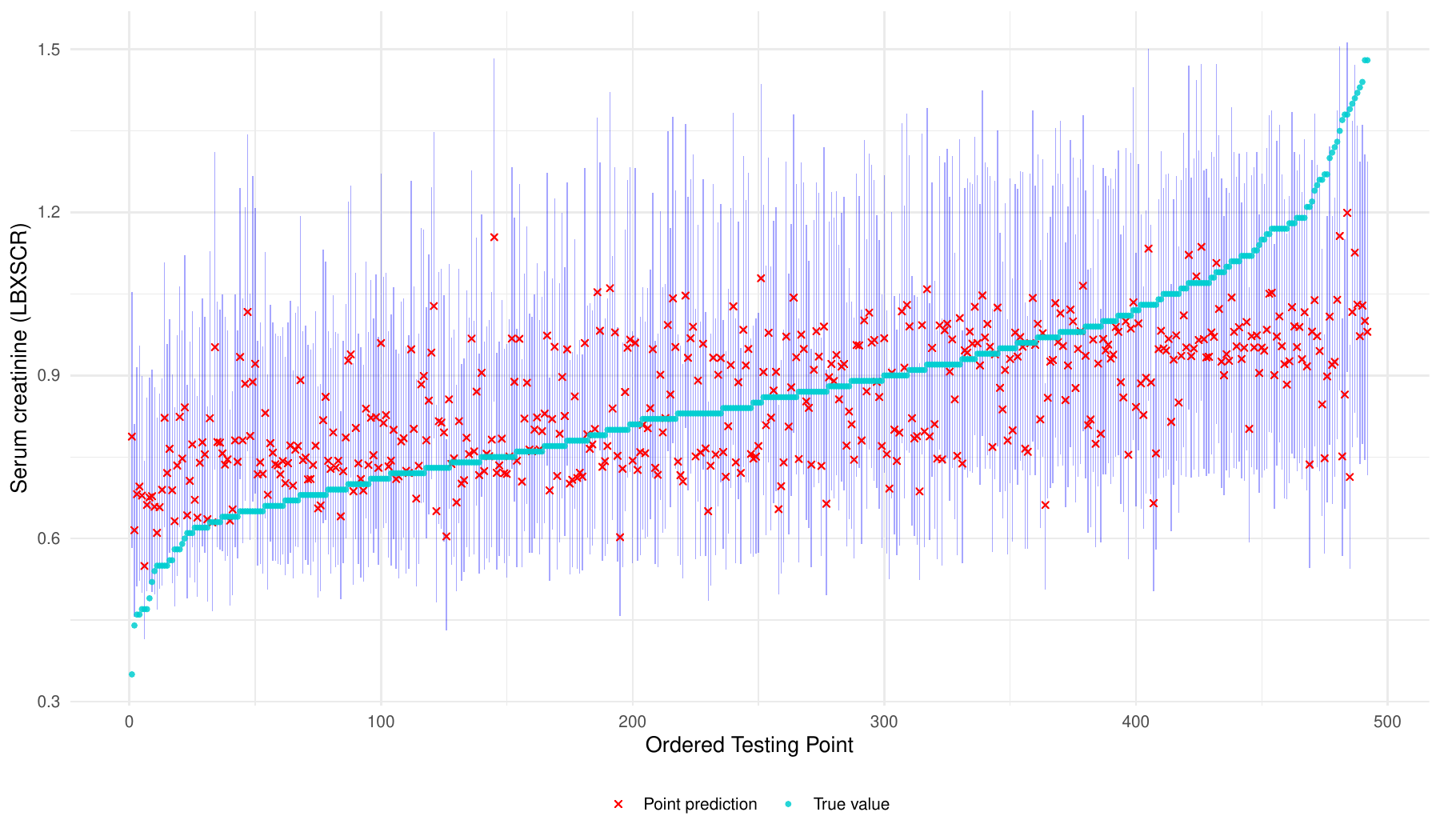}
  \caption{\textbf{Full-CSL prediction intervals for testing observations ordered by the observed response after removing outliers.}}
  \label{fig:testing_performance_removal}
\end{figure}

\begin{figure}[!htbp]
  \centering
  \includegraphics[width=0.9\textwidth]{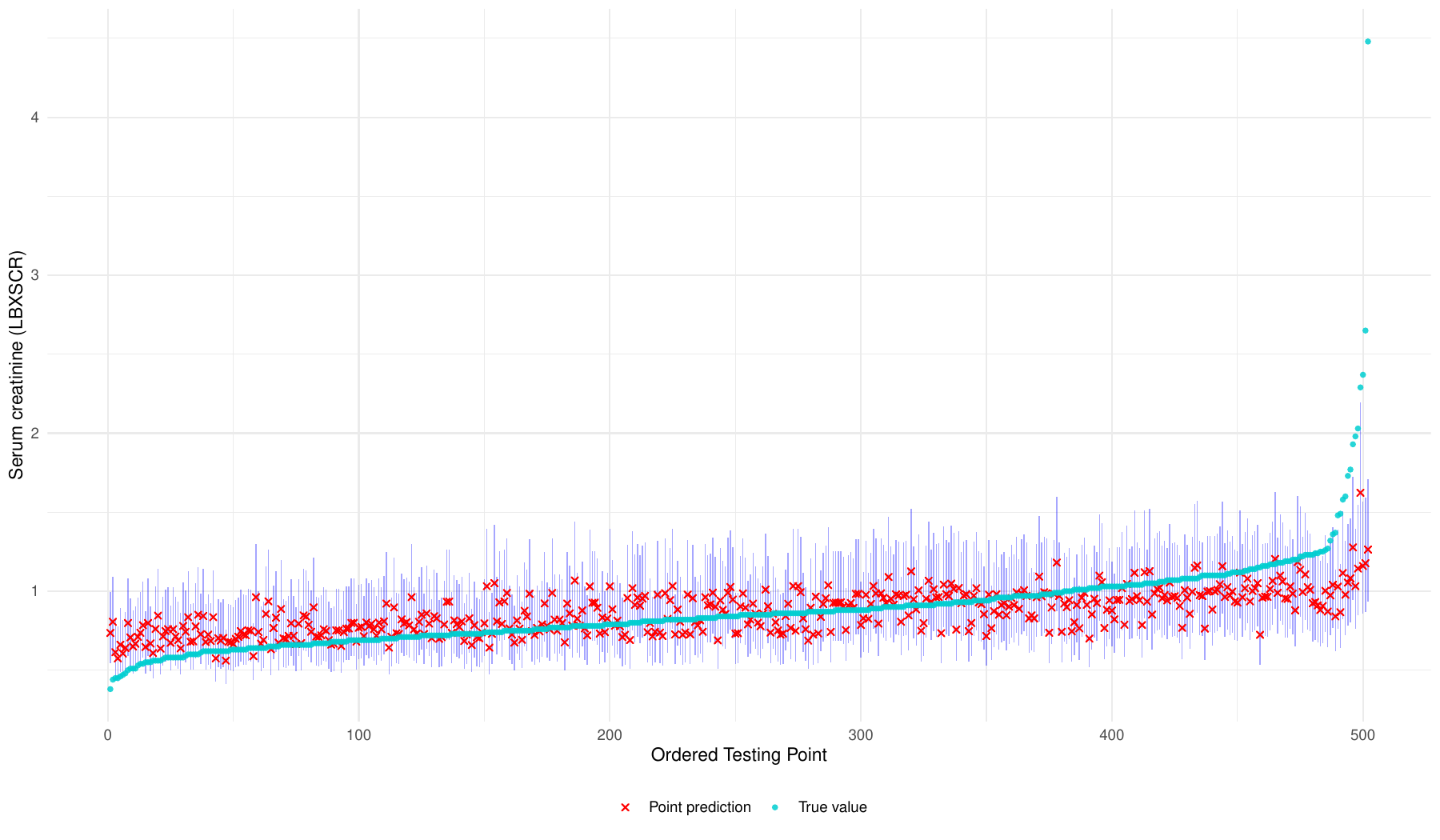}
  \caption{\textbf{Split-CSL prediction intervals for testing observations ordered by the observed response.}}
  \label{fig:split_testing_performance}
\end{figure}

\begin{figure}[!htbp]
  \centering
  \includegraphics[width=0.9\textwidth]{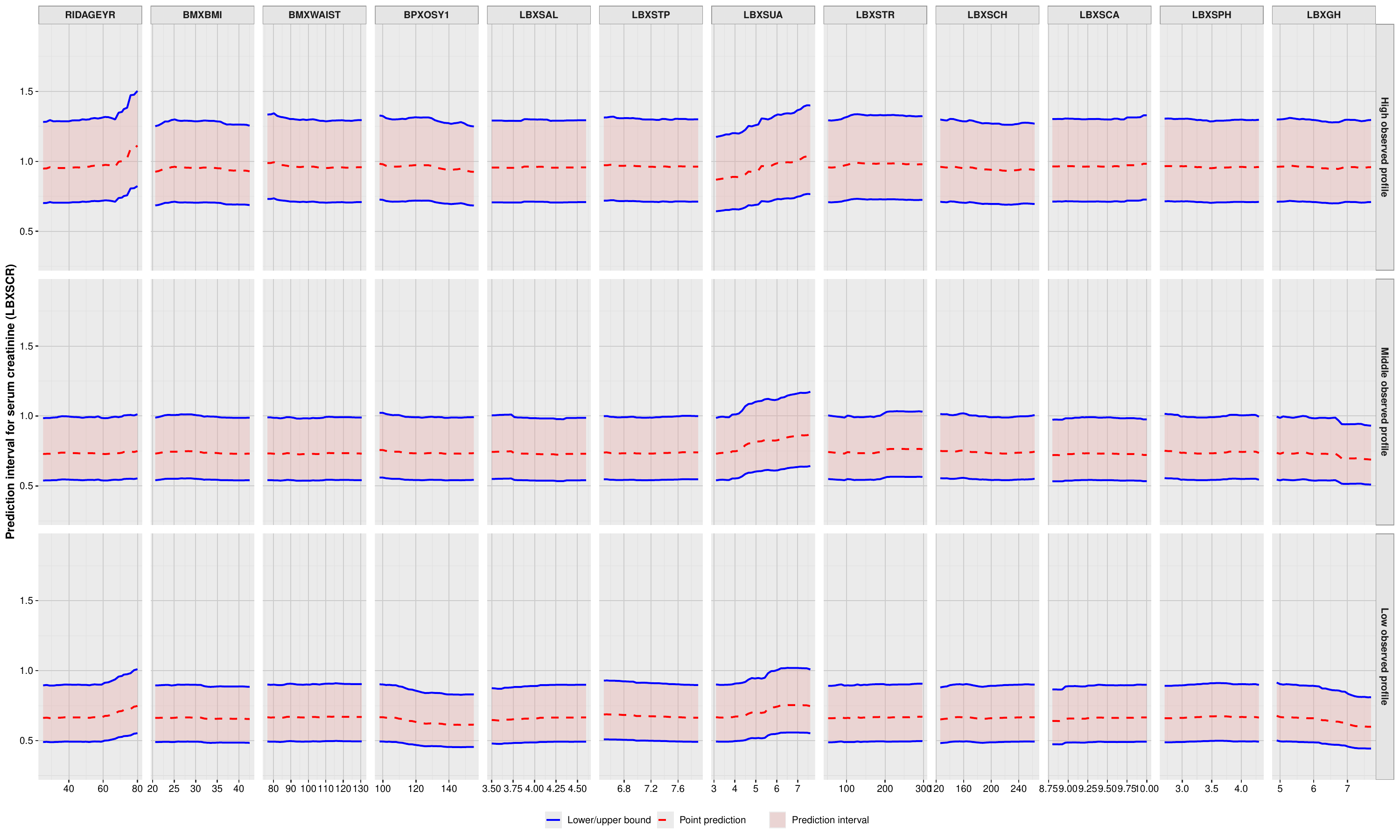}
  \caption{\textbf{Covariate-specific Split-CSL prediction intervals for three representative profiles.}}
  \label{fig:split_prediction_intervals_covariate}
\end{figure}

\begin{figure}[!htbp]
  \centering
  \includegraphics[width=0.9\textwidth]{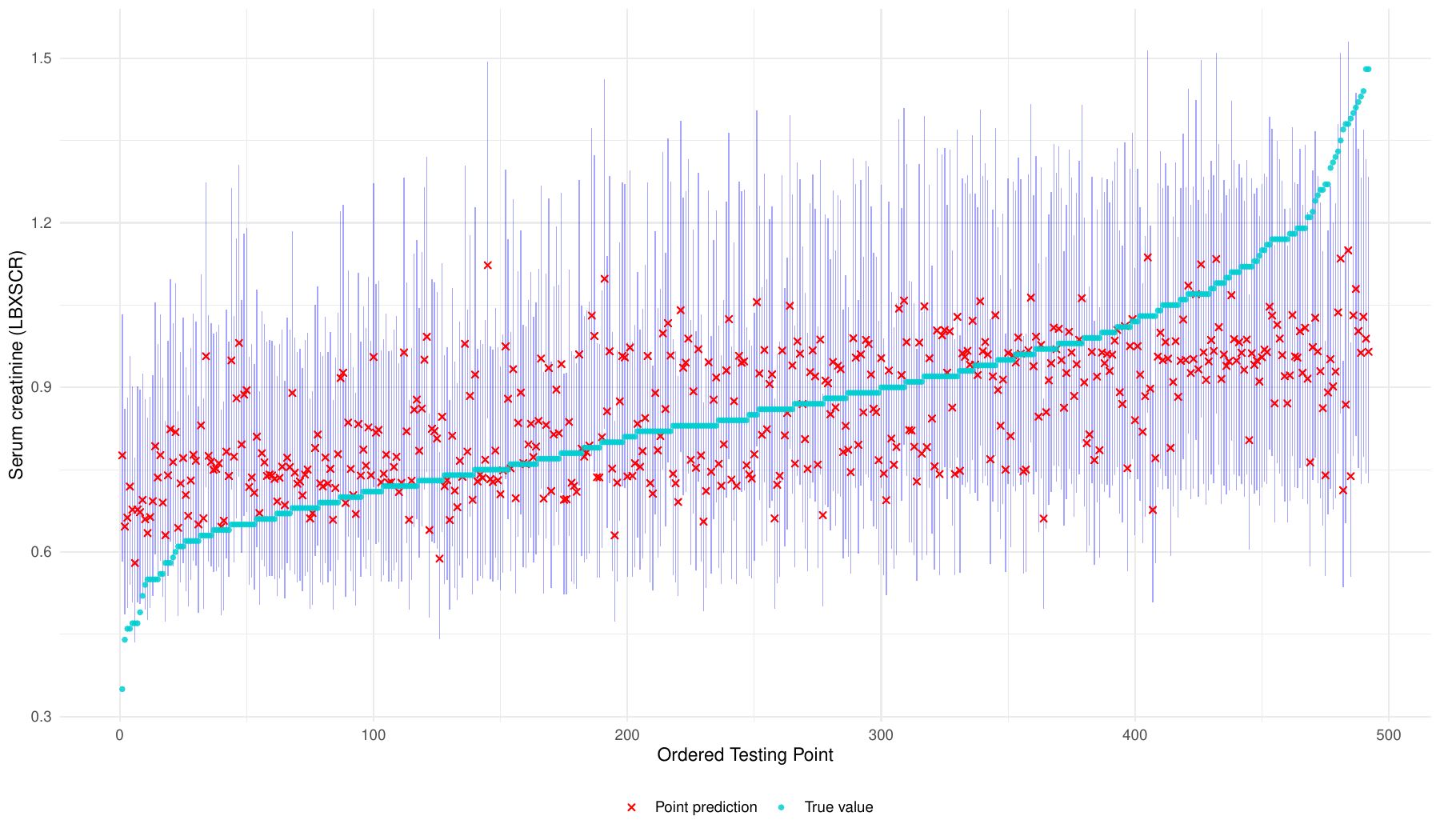}
  \caption{\textbf{Split-CSL prediction intervals for testing observations ordered by the observed response after removing outliers.}}
  \label{fig:split_testing_performance_removal}
\end{figure}

\begin{figure}[!htbp]
  \centering
  \includegraphics[width=0.9\textwidth]{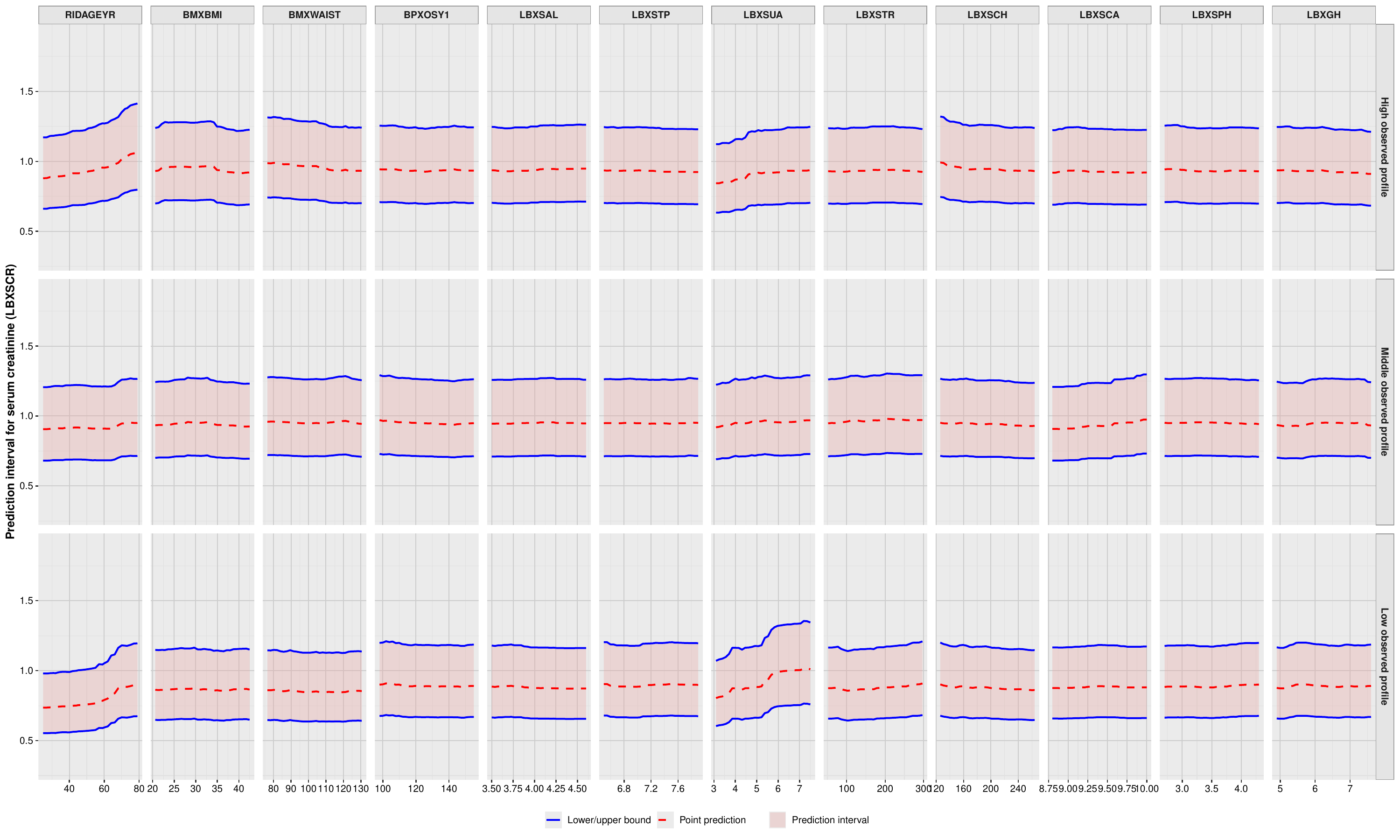}
  \caption{\textbf{Covariate-specific Split-CSL prediction intervals for three representative profiles after removing outliers.}}
  \label{fig:split_prediction_intervals_covariate_removal}
\end{figure}

\FloatBarrier

\subsubsection*{Sensitivity to split rules}

In split conformal prediction, the available data (excluding the testing set) must be partitioned into a training set and a calibration set. The training set is used to estimate the SL weights and fit each base learner, while the calibration set is used to compute the non-conformity scores and determine the conformal quantile thresholds. A larger training set improves the quality of the fitted models and the stability of the SL weight estimates, but leaves fewer observations for calibration, potentially leading to more variable quantile estimates and wider prediction intervals. Conversely, a larger calibration set yields more precise conformal thresholds but may lead to less stable weights and degrade the quality of the underlying predictors.

To assess the sensitivity of the Split-CSL procedure to this choice, we vary the training-to-calibration split ratio across six configurations. In the main case study analysis, we adopt an 80:20 split between training and calibration sets. Here, for the sensitivity check, we examine the following alternative ratios: 20:80, 30:70, 40:60, 50:50, 60:40, and 70:30. In each case, the test set remains fixed at 10\% of the total sample to ensure comparability across settings. We conduct this sensitivity analysis under both scenarios considered in the case study: the full data set ($n = 5027$) and the outlier-removed data set ($n = 4923$), where observations with serum creatinine exceeding 1.5 mg/dL are excluded. 

Tables \ref{tab:split_sensitivity_not} and \ref{tab:split_sensitivity_rem} report the estimated SL weights, empirical coverage, and mean interval width for each split ratio 
under the two scenarios. Several patterns can be observed. Firstly, all split ratios achieve empirical coverage of the nominal 90\% level in both scenarios. The mean interval widths are also similar across configurations, ranging from 0.5407 to 0.5593 in the full data set and from 0.4922 to 0.5056 in the outlier-removed data set. This suggests that the performance of the prediction intervals is not sensitive to the choice of split ratio. Secondly, the estimated weights change with the split ratio. In Table \ref{tab:split_sensitivity_not}, RF receives a dominant weight (exceeding 0.5) across all split ratios, though the margin is small at the 20:80 split rule, where weights are spread across LASSO and GAMLSS. As the training set becomes larger, the weights concentrate further on RF and LM, while LASSO and GAMLSS receive negligible weights. In Table \ref{tab:split_sensitivity_rem}, the pattern is different: LM receives a dominant weight only at the 20:80 split, RF becomes dominant only at the 70:30 split, and the remaining configurations have no dominant learner. This change reflects the reduced variability in the response after removing extreme values, which makes simpler models like LM more competitive with flexible learners such as RF, leading to more dispersed weight allocations.

\begin{table}[!htbp]
\centering
\small
\begin{tabular}{l*{6}{c}c}
\toprule
& \multicolumn{6}{c}{\textbf{SL Weights}} & \\
\cmidrule(lr){2-7}
\textbf{Split (Train:Calib)} & \textbf{LM} & \textbf{LASSO} & \textbf{GAM} & \textbf{NNET} & \textbf{RF} & \textbf{GAMLSS} & \textbf{Coverage/Mean Width} \\
\midrule
\textbf{20:80} & 0.0328 & 0.1934 & 0.0000 & 0.0225 & 0.5001 & 0.2512 & 0.910/0.5593  \\
\textbf{30:70} & 0.0601 & 0.1206 & 0.0656 & 0.0000 & 0.5864 & 0.1673 & 0.910/0.5500  \\
\textbf{40:60} & 0.3394 & 0.0000 & 0.0620 & 0.0000 & 0.5303 & 0.0683 & 0.910/0.5498  \\
\textbf{50:50} & 0.1500 & 0.0000 & 0.3138 & 0.0000 & 0.5245 & 0.0117 & 0.906/0.5408  \\
\textbf{60:40} & 0.3196 & 0.0000 & 0.1213 & 0.0000 & 0.5591 & 0.0000 & 0.918/0.5499  \\
\textbf{70:30} & 0.3397 & 0.0000 & 0.0807 & 0.0000 & 0.5796 & 0.0000 & 0.912/0.5407  \\
\bottomrule
\end{tabular}
\caption{\textbf{Split-CSL performance under different split ratios without removing outliers.}}
\label{tab:split_sensitivity_not}
\end{table}

\begin{table}[!htbp]
\centering
\small
\begin{tabular}{l*{6}{c}c}
\toprule
& \multicolumn{6}{c}{\textbf{SL Weights}} & \\
\cmidrule(lr){2-7}
\textbf{Split (Train:Calib)} & \textbf{LM} & \textbf{LASSO} & \textbf{GAM} & \textbf{NNET} & \textbf{RF} & \textbf{GAMLSS} & \textbf{Coverage/Mean Width} \\
\midrule
\textbf{20:80} & 0.5264 & 0.0000 & 0.0000 & 0.0730 & 0.3076 & 0.0930 & 0.911/0.5056  \\
\textbf{30:70} & 0.3048 & 0.0000 & 0.0583 & 0.0000 & 0.4398 & 0.1971 & 0.913/0.4932  \\
\textbf{40:60} & 0.3861 & 0.0000 & 0.1753 & 0.0110 & 0.3714 & 0.0562 & 0.921/0.4970  \\
\textbf{50:50} & 0.1129 & 0.2880 & 0.1203 & 0.0000 & 0.4788 & 0.0000 & 0.909/0.4922  \\
\textbf{60:40} & 0.4407 & 0.0000 & 0.0516 & 0.0204 & 0.4873 & 0.0000 & 0.917/0.4922  \\
\textbf{70:30} & 0.3529 & 0.0000 & 0.0516 & 0.0000 & 0.5955 & 0.0000 & 0.909/0.4937  \\
\bottomrule
\end{tabular}
\caption{\textbf{Split-CSL performance under different split ratios with outliers removed.}}
\label{tab:split_sensitivity_rem}
\end{table}

\subsubsection*{Categorical variable analysis}

\begin{figure}[!htbp]
  \centering
  \includegraphics[width=0.7\textwidth]{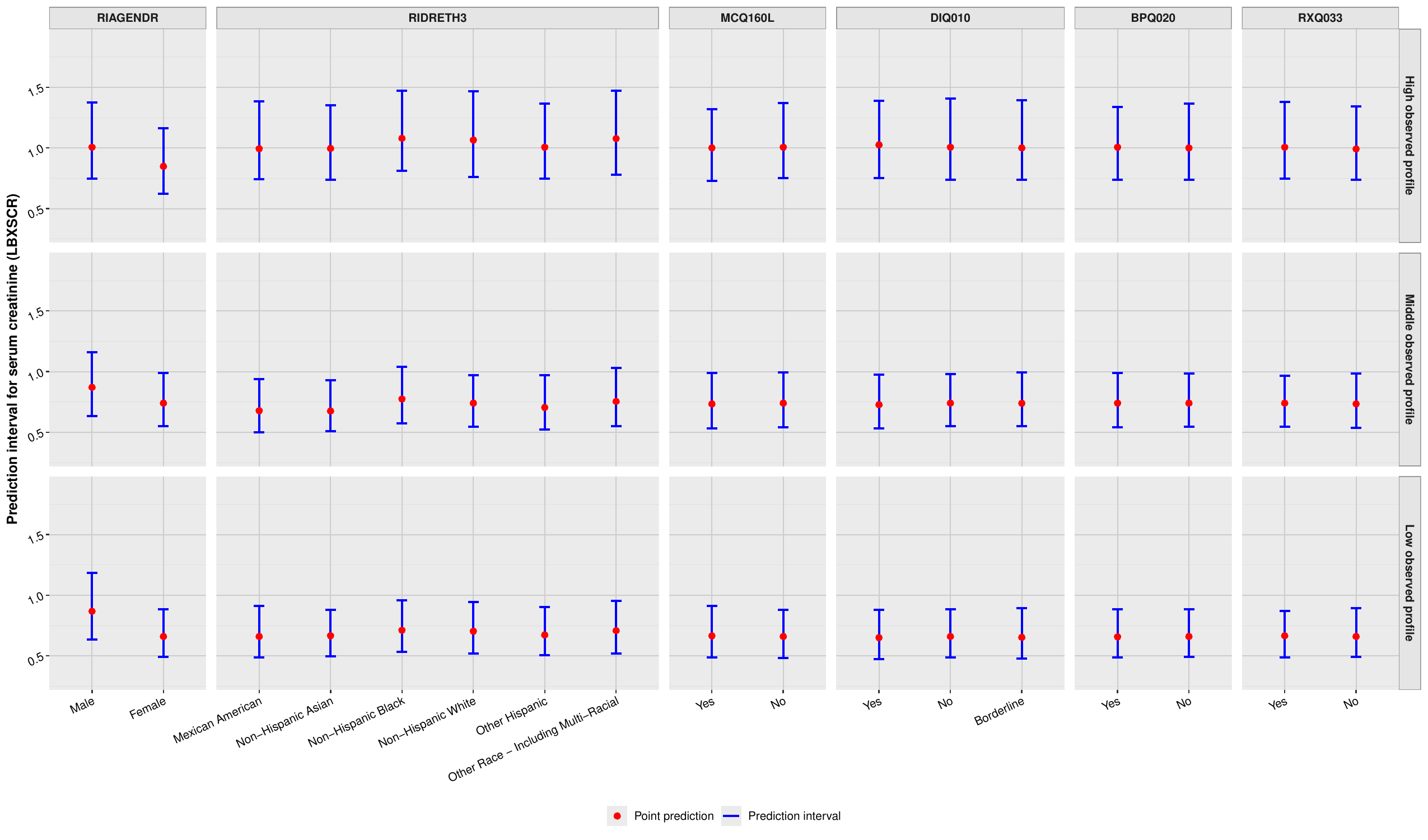}
  \caption{\textbf{Categorical covariate-specific Full-CSL prediction intervals for three representative profiles.}}
  \label{fig:categorical_full}
\end{figure}

\begin{figure}[!htbp]
  \centering
  \includegraphics[width=0.7\textwidth]{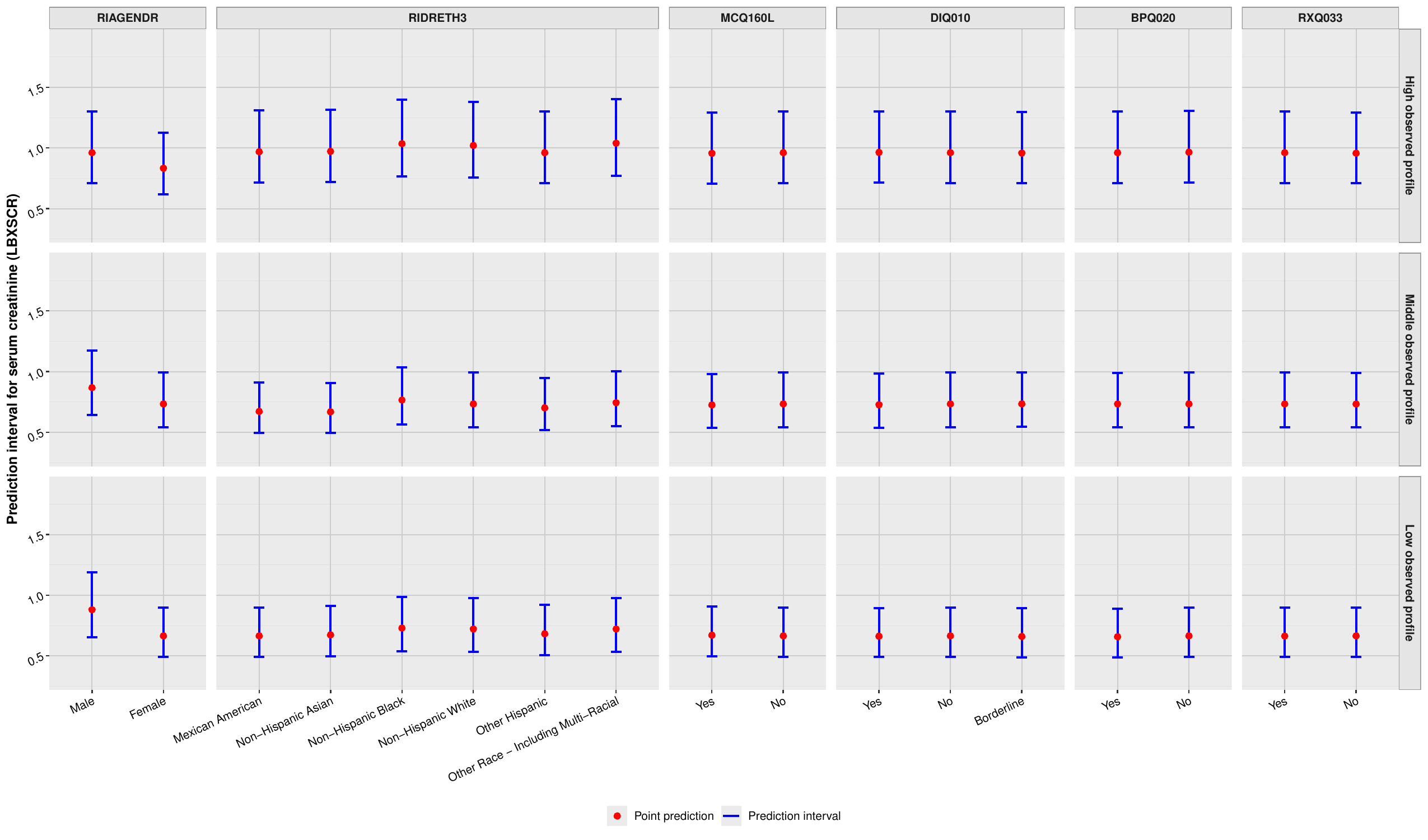}
  \caption{\textbf{Categorical covariate-specific Split-CSL prediction intervals for three representative profiles.}}
  \label{fig:categorical_split}
\end{figure}

\begin{figure}[!htbp]
  \centering
  \includegraphics[width=0.7\textwidth]{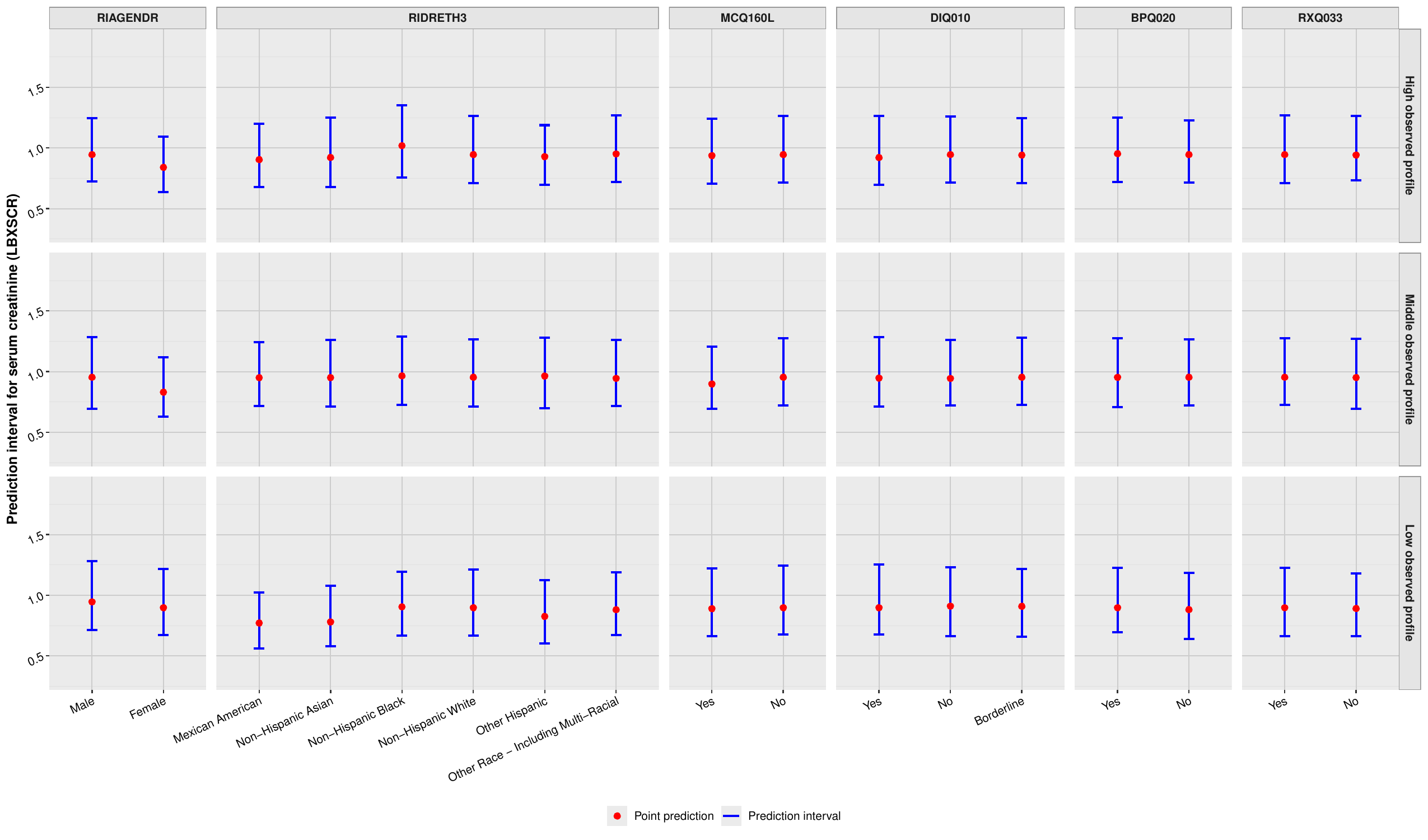}
  \caption{\textbf{Categorical covariate-specific Full-CSL prediction intervals for three representative profiles after removing outliers.}}
  \label{fig:categorical_full_removal}
\end{figure}

\begin{figure}[!htbp]
  \centering
  \includegraphics[width=0.7\textwidth]{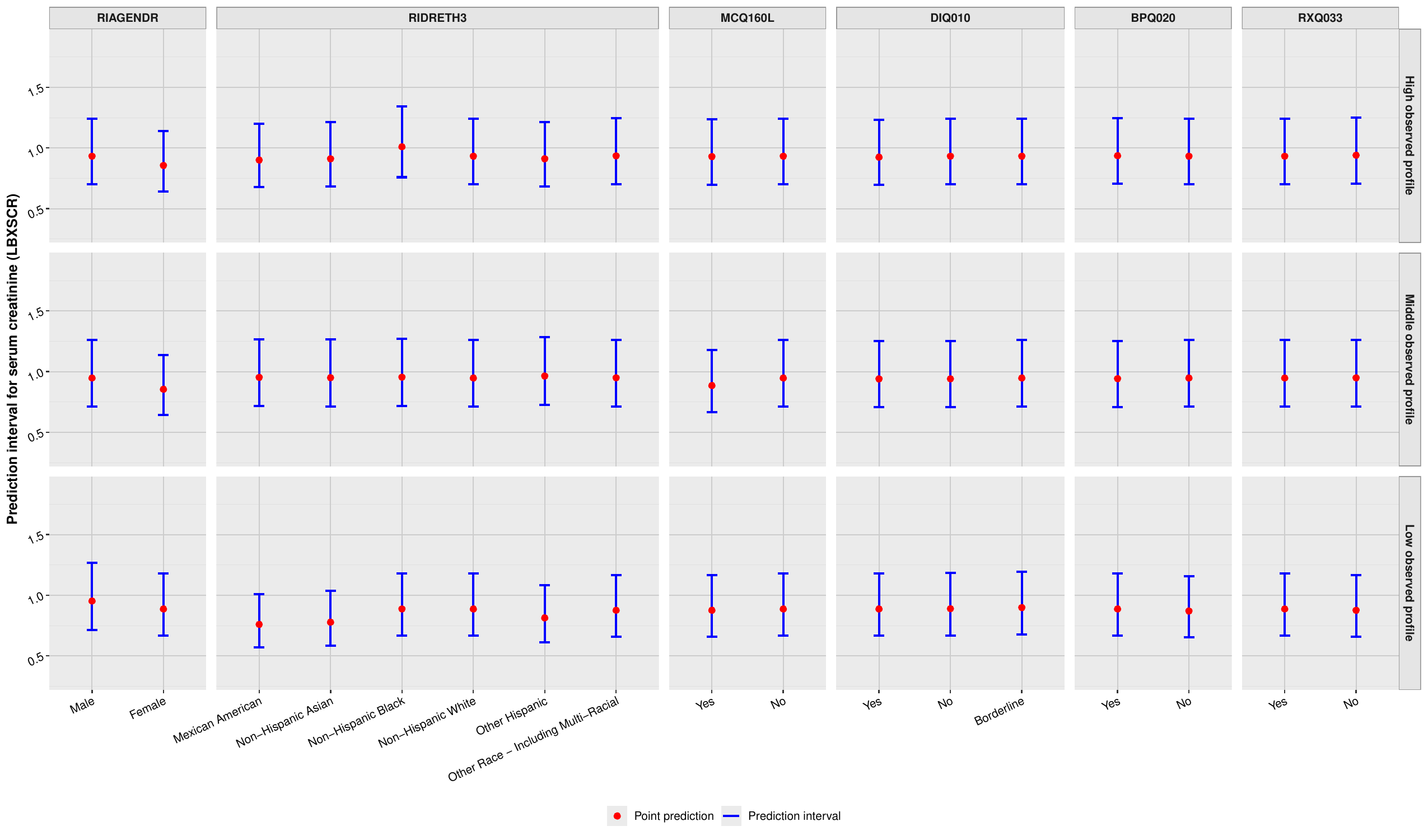}
  \caption{\textbf{Categorical covariate-specific Split-CSL prediction intervals for three representative profiles after removing outliers.}}
  \label{fig:categorical_split_removal}
\end{figure}

Figures \ref{fig:categorical_full}, \ref{fig:categorical_split}, \ref{fig:categorical_full_removal}, and \ref{fig:categorical_split_removal} extend the covariate-specific analysis to the six categorical covariates considered in the case study: sex (\texttt{RIAGENDR}), race/ethnicity (\texttt{RIDRETH3}), liver disease status (\texttt{MCQ160L}), diabetes status (\texttt{DIQ010}), high blood pressure status (\texttt{BPQ020}), and prescription medication use (\texttt{RXQ033}). The same three representative profiles are analyzed under both Full-CSL and Split-CSL, with and without removing extreme observations. Sex shows the clearest and most consistent effect: both the point prediction and the prediction interval for serum creatinine is systematically higher for males than for females in all three profiles, and this difference is present under both conformal procedures and remains after outlier removal. Race/ethnicity exhibits only modest heterogeneity, with some small differences in point predictions across groups, but without a stable ordering across profiles and conformal settings. By contrast, the liver disease status, diabetes status, high blood pressure status, and prescription medication use produce largely overlapping point predictions and interval bounds, indicating limited marginal influence on predicted creatinine once the remaining covariates are held fixed. Overall, the conclusions are qualitatively similar for Full-CSL and Split-CSL. After removing outliers, the intervals become slightly narrower and the differences between categories are milder, but the main pattern remains unchanged.

\end{document}